\documentclass[lettersize,journal]{IEEEtran}
\usepackage{amsmath,amsfonts,amsthm}
\usepackage{algorithmic}
\usepackage{algorithm}
\usepackage{array}
\usepackage[caption=false,font=normalsize,labelfont=sf,textfont=sf]{subfig}
\usepackage{textcomp}
\usepackage{stfloats}
\usepackage{url}
\usepackage{verbatim}
\usepackage{graphicx}
\usepackage{cite}
\usepackage[short]{optidef}
\usepackage{longtable} 
\usepackage{booktabs}
\usepackage{adjustbox}
\usepackage[dvipsnames]{xcolor}
\usepackage{cite}
\usepackage[hidelinks]{hyperref}
\newcommand{\ra}[1]{\renewcommand{\arraystretch}{#1}} 

\newtheorem{definition}{Definition}[section] 
\newtheorem{property}{Property}[section] 
\newtheorem{Assumption}{Assumption}[section]

\usepackage{tikz}
\newcommand\copyrighttext{%
	\footnotesize \textcopyright 2024 IEEE. Personal use of this material is permitted. Permission from IEEE must be obtained for all other uses, in any current or future media, including reprinting/republishing this material for advertising or promotional purposes, creating new collective works, for resale or redistribution to servers or lists, or reuse of any copyrighted component of this work in other works.}
\newcommand\copyrightnotice{%
	\begin{tikzpicture}[remember picture,overlay]
		\node[anchor=south,yshift=5pt] at (current page.south) {\fbox{\parbox{\dimexpr\textwidth-\fboxsep-\fboxrule\relax}{\copyrighttext}}};
	\end{tikzpicture}%
}

\usepackage{acronym}
\newacro{QP}{quadratic program}
\newacro{MIQP}{mixed-integer quadratic program}
\newacro{MINLP}{mixed-integer nonlinear programming}
\newacro{MIOCP}{mixed-integer optimal control problems}
\newacro{MI}{mixed-integer}
\newacro{MIP}{mixed-integer program}
\newacro{NLP}{nonlinear program}
\newacro{NMPC}{nonlinear model predictive controller}
\newacro{EDS}{permutation equivariant deep set}
\newacro{REDS}{recurrent permutation equivariant deep set}
\newacro{FF}{feed forward}
\newacro{DNN}{deep neural network}
\newacro{FCF}{Frenet coordinate frame}
\newacro{CCF}{Cartesian coordinate frame}
\newacro{FP}{feasibility projector}
\newacro{SQP}{sequential quadratic programming}
\newacro{RL}{reinforcement learning}
\newacro{COCO}{\emph{combinatorial offline convex online}}
\newacro{CC}{Chebychev center}
\newacro{LSTMP}{long-short-term motion planner}
\newacro{MIP-DM}{mixed-integer programming-based vehicle decision maker}
\newacro{STF}{short-term motion planning formulation}
\newacro{LTF}{long-term motion planning formulation}
\newacro{OCP}{optimal control problem}
\newacro{EV}{ego vehicle}
\newacro{SV}{surrounding vehicle}
\newacro{MIP-DM}{mixed-integer motion planning and decision maker}
\newacro{AD}{autonomous driving}
\newacro{ST}{spatio-temporal}
\newacro{SLT}{spatio-lateral-temporal}
\newacro{VT}{velocity-time}
\newacro{ODE}{ordinary differential equation}
\newacro{AI}{artificial intelligence}
\newacro{NN}{neural network}
\newacro{RL}{reinforcement learning}

\newcommand{\st}{\ac{ST}-space}
\newcommand{\slt}{\ac{SLT}-space}

\newcommand{\occspace}{\mathcal{O}}

\newcommand{\latspace}[1]{\mathcal{N}_{#1}}

\newcommand{\freeSpace}{\mathcal{F}}
\newcommand{\freeSpaceTwoDim}{\mathcal{S}}

\newcommand{\reachspace}[1]{\mathcal{R}_{#1}}
\newcommand{\reachspaceTrue}[1]{\mathcal{R}^\mathrm{true}_{#1}}

\newcommand{\stateAdmiss}{\mathcal{X}}
\newcommand{\stateFree}{\mathcal{X}_\mathrm{free}}
\newcommand{\conrlAdmiss}{\mathcal{U}}

\newcommand{\slack}{q}
\newcommand{\Slack}{Q}

\newcommand{\astar}{hybrid $\mathrm{A}^*$}

\newcommand{\sumo}{\texttt{sumo}}

\newcommand{\gurobi}{\texttt{gurobi}}

\newcommand{\shortHorPlanner}{\ac{STF}}
\newcommand{\longHorPlanner}{\ac{LTF}}
\newcommand{\lstp}{\ac{LSTMP}}

\newcommand{\opp}{\ac{SV}}
\newcommand{\opps}{\acp{SV}}

\newcommand{\nmpc}{low-level \ac{NMPC}}

\newcommand{\refer}[1]{\tilde #1}
\newcommand{\ub}[1]{\overline{#1}}
\newcommand{\lb}[1]{\underline{#1}}

\newcommand{\nHor}{N}
\newcommand{\tHor}{t_\mathrm{f}}

\newcommand{\td}{t_\mathrm{d}}

\newcommand{\dimT}{t_\mathrm{f}}

\newcommand{\dimLanes}{L}

\newcommand{\laneWidth}{d_\mathrm{lane}}

\newcommand{\R}{\mathbb{R}}
\newcommand{\Z}{\mathbb{N}}

\newcommand{\norm}[1]{\left\lVert #1 \right\rVert}
\newcommand{\intSet}[2][1]{\mathbb{N}_{[{#1}:{#2}]}}

\newcommand{\minus}{\scalebox{0.75}[1.0]{$-$}}

\begin{document}

\title{A Long-Short-Term Mixed-Integer Formulation for Highway Lane Change Planning}

\author{Rudolf Reiter,
	Armin Nurkanovi\'c,
	Daniele Bernardini,
	Moritz Diehl,
 	Alberto Bemporad,~\IEEEmembership{Fellow,~IEEE}
        % <-this % stops a space
\thanks{R. Reiter, A. Nurkanovi\'c and M. Diehl are with 
	the Unversity of Freiburg, 79110 Freiburg i. B., Germany (e-mails: \{rudolf.reiter,armin.nurkanovic,moritz.diehl\}@imtek.uni-freiburg.com).}
\thanks{D. Bernardini is with 
	ODYS S.r.l., 20159 Milano, Italy (e-mail: daniele.bernardini@odys.it).}
\thanks{A. Bemporad is with the IMT School for Advanced Studies
	Lucca, 55100 Lucca, Italy (e-mail: alberto.bemporad@imtlucca.it).}
\thanks{This work was supported by EU via ELO-X 953348, by DFG via Research Unit FOR 2401, project 424107692 and 525018088 and by BMWK via 03EI4057A and 03EN3054B.}% <-this % stops a space
}

%{Reiter \MakeLowercase{\textit{et al.}}: A Long-Short-Term Mixed-Integer Formulation for Highway Lane Change Planning}


\maketitle
\copyrightnotice

\begin{abstract}
This work considers the problem of optimal lane changing in a structured multi-agent road environment. 
A novel motion planning algorithm that can capture long-horizon dependencies as well as short-horizon dynamics is presented. 
Pivotal to our approach is a geometric approximation of the long-horizon combinatorial transition problem which we formulate in the continuous time-space domain.
Moreover, a discrete-time formulation of a short-horizon optimal motion planning problem is formulated and combined with the long-horizon planner.
Both individual problems, as well as their combination, are formulated as \acp{MIQP} and solved in real-time by using state-of-the-art solvers.
We show how the presented algorithm outperforms two other state-of-the-art motion planning algorithms in closed-loop performance and computation time in lane changing problems.
Evaluations are performed using the traffic simulator \texttt{SUMO}, a custom low-level tracking model predictive controller, and high-fidelity vehicle models and scenarios, provided by the \texttt{CommonRoad} environment.
\end{abstract}


\section{Introduction}
\IEEEPARstart{I}{n} recent years many approaches have been proposed for vehicle motion planning in structured multi-lane road environments. However, considering combinatorial long-term dependencies and providing optimal trajectories subject to dynamic constraints in real-time remains a challenging problem.
In fact, even deterministic two-dimensional motion planning problems with rectangular obstacles are NP-hard~\cite{Reif1979, LaValle2006}.

\begin{figure}
	\begin{center}
		\includegraphics[scale=1]{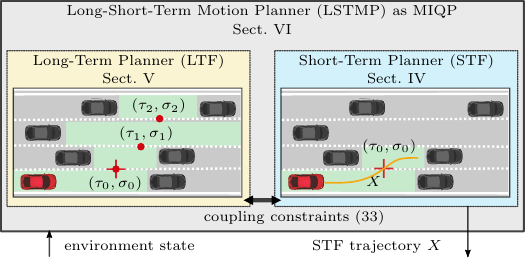}
		\caption{Overview of the proposed \ac{MIQP} formulation for motion planning, referred to as \ac{LSTMP}. The \ac{MIQP} consists of long-term and short-term planning formulations where the decision variables of both are coupled through \emph{consistency} constraints. The short-term decision variables include a continuous point-mass model trajectory to approximate a single lane change. The long-term decision variables account for selecting gaps between \acp{SV} on each lane.}
		\label{fig:overview}
	\end{center}
\end{figure}

This work proposes a novel iterative planning algorithm, referred to as \lstp{} that reduces the combinatorial complexity by splitting the problem into a \shortHorPlanner{} and a \longHorPlanner{}, both solved by one \ac{MIQP}, cf. Fig.~\ref{fig:overview}.
The \shortHorPlanner{} aims at optimizing a four-state discrete-time trajectory of a point-mass model including obstacle constraints, similar to the formulations of~\cite{Quirynen2023,Miller2018}. The \shortHorPlanner{} trajectory is computed for a shorter horizon to approximate a maximum of one lane change. In contrast, the \longHorPlanner{} aims at obtaining optimal lane transitions, defined by the transition times and longitudinal transition positions, which are both continuous variables. These lane transitions are used for long-term planning, i.e., the choice of gaps between vehicles on several consecutive lanes. Reachability and the choice of transition gaps on consecutive lanes are modeled by disjunctive programming. 

The planned trajectory of the \shortHorPlanner{} and the transitions of the \longHorPlanner{} are formulated consistently, i.e., a transition point constrains the point-mass model trajectory to the corresponding lane. 
Contrary to strict hierarchical decomposition, the coarser approximation of the high-level plan cannot be infeasible for the low-level planner.

A challenge of state-of-the-art motion planners is the scaling of computational complexity with the horizon length~\cite{Quirynen2023}, which makes long-horizon planning most often intractable. 
Within the formulation of the \longHorPlanner{}, the locations of transitions in time and position are continuous. The proposed modeling uses integer variables only related to the gaps between vehicles on each lane. Therefore, the number of integer variables does not scale with the horizon length within the \longHorPlanner{}. Consequently, the constant small number of integer variables, even for long-term predictions, allows for fast computation times of the algorithm.

Evaluations of the proposed approach are performed with both deterministic and interactive closed-loop simulations that involve a \texttt{CommonRoad}~\cite{Althoff2017} vehicle model, a low-level \ac{NMPC}, and interactive traffic that is simulated with \sumo{}.\\\\

\subsection{Related work}
An abundance of fundamentally different techniques address highway motion planning and were reviewed by~\cite{Paden2016} and, more recently, by~\cite{Claussmann2020,Reda2024} and particularly for deep learning in~\cite{Szilard2022, Ravi2022}. The authors in~\cite{Paden2016} identified geometric, variational, graph-search, and incremental search methods as fundamental planning categories, whereas the more recent and exhaustive survey~\cite{Claussmann2020} adds a major part on \ac{AI}, among further refinements.

Historically, geometric and rule-based approaches were more dominant.
Parametric curves are used in highly structured environments, such as highways, due to their simplicity and ease of alignment with the road geometry~\cite{Mouhagir2016,Sheckells2017, Klancar2021}. However, the motion plans are usually conservative and unable to cope with complex environments~\cite{Claussmann2020}.

In several works, the state space is discretized and some sort of graph search is performed~\cite{Paden2016,Reda2024}.
Probabilistic road maps~\cite{Kavraki1996,Karaman2011}, Djikstra graph search and similar, rapidly exploring random trees~\cite{Oktay2017,Wang2020a,Wang2021} are common in highway motion planning. Nonetheless, they suffer from the curse of dimensionality, the inability to handle dynamics, a poor connectivity graph, and a poor repeatability of results~\cite{Claussmann2020}.
By using heuristics, \astar{}~\cite{Montemerlo2008, Ajanovic2018,Dongchan2022} aims to avoid the problem of high-dimensional discretization in graph-search. Choosing an admissible heuristic is challenging and a time-consuming graph generation is performed in each iteration. 
Moreover, authors try to improve graph- or sampling-based algorithms by combining them with learning-based methods~\cite{Wang2020a, Jikun2020}.

Optimization-based methods can successfully solve motion planning problems in high-dimensional state spaces in real-time~\cite{Diehl2005}. They are appealing due to numerous advantages, e.g., consideration of dynamics and constraints, adaptability to new scenarios, finding and keeping solutions despite environmental changes, and taking into account complex scenarios.
Pure derivative-based methods are often restricted to convex problem structures or sufficiently good initial guesses.
Highway motion planning is highly non-convex. Nonetheless, by introducing integer variables, the problem can be formulated as an \ac{MIQP}~\cite{Xiangjun2016,Miller2018,Li2022,Quirynen2023} and solved by dedicated high-performance solvers, such as \gurobi{}~\cite{Gurobi2023}.
Yet, the planning horizon of \ac{MIQP} with a fixed discrete-time trajectory is still limited due to the increasing number of integer variables for increased horizon lengths. Therefore, keeping the computation time limited remains a challenge,~\cite{Claussmann2020}.

One successful structure exploitation for solving the highway motion planning problem faster is the decomposition of the state space into \emph{spatio-temporal} driving corridors~\cite{Johnson2013,Bender2015,Miller2018,Marcucci2022,Li2022, Deolasee23} with \emph{simple} obstacle predictions.
Still highly non-convex, the region can be decomposed into convex cells~\cite{Marcucci2022,Deolasee23} or used in a sampling-based planner~\cite{Li2022}.

To leverage the computational burden further, without sacrificing significantly the overall performance, the presented approach uses a short-horizon planning similar to~\cite{Quirynen2023} and~\cite{Miller2018}, adding a long-horizon coarse geometric approximation in the spatio-temporal domain (see Fig.~\ref{fig:overview}). 
The proposed \shortHorPlanner{} differs from~\cite{Miller2018} by using only one binary variable per time step for the first lane change and more accurately modeling occupied regions, that also consider braking due to preceding obstacles. Moreover, the consecutive gap is not fixed as in~\cite{Miller2018} but determined by the \longHorPlanner{}.
The idea of combining two horizons was presented in~\cite{Laurense2022}, yet not related to combinatorial motion planning and hierarchically decomposed in~\cite{Dongchan2022} using a graph-based planner. 

\ac{AI}-based methods often use exhaustive simulations to train \acp{NN} by \ac{RL}~\cite{Ravi2022} or use expert data, such as data collected from human divers, to perform imitation learning~\cite{Wang2021a,Bronstein2022}. They often struggle to consider safety critical constraints and adapt to environment changes~\cite{Claussmann2020,Szilard2022, Bronstein2022}. Furthermore, sim-to-real challenges apply~\cite{Szalay2018} since these methods are mainly trained within simulations. An advantage includes the capability of using raw sensor inputs, such as camera images and the low computational requirements of trained \acp{NN}~\cite{Szilard2022}. 

The performance of the \lstp{} is compared to a state-of-the-art \astar{} method~\cite{Ajanovic2018}, which can be classified as both a deterministic planning \ac{AI} and graph-search method~\cite{Russell2020}, and to the \ac{MIP-DM}~\cite{Quirynen2023} which is a comparable state-of-the-art optimization-based method.
 
\subsection{Contribution}
This paper contributes a novel algorithm for optimal lane-changing highway maneuvers.
Compared to other highway motion planning algorithms, the proposed lane change motion planner approximates long-term dependencies in the \st{}, where the computational burden is independent of the position and the time a lane change occurs.
Moreover, we solve the problem involving the long-term approximation as a consistent single problem and, therefore, avoid a problematic decoupling.
The closed-loop performance is improved by~$15\%$ compared to~\cite{Quirynen2023} and~\cite{Ajanovic2018} and the average computation time is lowered in randomized interactive simulations by two orders of magnitude. Compared to~\cite{Quirynen2023}, the the number of integer variables used within the underlying optimization problem reduces from $O(N_\mathrm{veh}N)$ to~$O(N_\mathrm{veh}+N)$ for a number of~$N_\mathrm{veh}$ \acp{SV} and $N$ discrete-time prediction steps, with a comparable closed-loop performance on the evaluated scenarios.

\subsection{Outline}
In Sect.~\ref{sec:preliminary}, important background concepts are defined that are used throughout the paper. In Sect.~\ref{sec:general_problem}, the general problem definition, related assumptions, and simplifications are introduced. Next, the planning approach is described in Sect.~\ref{sec:short_planner} to~\ref{sec:long_short_planner} and evaluated in Sect.~\ref{sec:evaluation}. The approach is discussed and conclusions are drawn in Sect.~\ref{sec:conclusion}.

\section{Preliminaries and Notation} 
\label{sec:preliminary}
The set of non-negative real numbers is denoted by~$\R^+=\{x\in\R\mid x\geq0\}$ and non-negative integers by~$\mathbb{N}_0=\{x\in\mathbb{Z}\mid x\geq0\}$.
Integer sets are written as~$\intSet[m]{n}=\{z \in \mathbb{N}_0 \mid m \leq z \leq n\}$ with $m< n$.
By using the notation~$f(x;y)$ in the context of optimization problems, we denote the dependency of function~$f$ on variables~$x$ and parameters~$y$.
The convex hull~$C\subseteq\R^n$ of two polygons~$A\subseteq\R^n$ and~$B\subseteq\R^n$ is $C=\mathrm{conv}(A,B)$.
We use the floor function~$\lfloor x\rfloor$ for rounding a number~$x\in\R$ to the largest smaller integer and the ceil function~$\lceil x\rceil$ for rounding to the smallest larger integer.

\subsection{Propositional logic and mixed-integer notation}
For a given compact set~$\mathcal{X}\subset \R$ and continuous function~$f:\mathcal{X}\rightarrow\R$, let~$\ub{M}\geq\max_{x\in \mathcal{X}}f(x)$ and~$\lb{M}\leq\min_{x\in \mathcal{X}}f(x)$ denote an upper and lower bound of $f(x)$ on $\mathcal{X}$, respectively.
The following properties hold for a given~$f:\mathcal{X}\rightarrow\R$~\cite{Torrisi2004, Williams2013}.
\begin{property}
	\label{def:bilinear}
	For a product~$y=\beta f(x)$, with~$y\in\R$, the following equivalence holds for all~$x\in\mathcal{X}$ and~$\beta \in \intSet[0]{1}$:
	\begin{align}
		y = \beta f(x) \Leftrightarrow 
		\begin{cases}
		y &\leq \ub{M} \beta ,\\
		y &\geq \lb{M} \beta, \\
		y &\leq f(x) - \lb{M}(1-\beta),\\
		y &\geq f(x) - \ub{M}(1-\beta).
		\end{cases}
	\end{align}
\end{property}
\begin{property}
	\label{def:bigm_activation}
	The implication $[\beta=1] \implies [f(x)\geq 0]$ of a binary variable $\beta \in \intSet[0]{1}$ that activates constraint~$f(x)\geq 0$, is formulated as
	\begin{equation}
		f(x)\geq \lb{M} (1-\beta).
	\end{equation}
\end{property}
\begin{property}
	\label{def:bigm_indicaiton}
	The implication $[f(x)> 0] \implies [\beta=1]$ of a binary variable~$\beta \in \intSet[0]{1}$ that gets activated if constraint~$f(x)> 0$ is valid, is formulated as
	\begin{equation}
		f(x)\leq  \ub{M} \beta.
	\end{equation}
\end{property}
\begin{property}
	\label{def:disjunction}
	The disjunction $\bigvee_{i=1}^{N} [f_i(x)\geq0]$ is formulated by adding~$N$ binary variables $\beta_i\in \intSet[0]{1}$ with $i\in \intSet{N}$, and the conditions
	\begin{align}
		\begin{split}
		[\beta_i=1]& \implies [f_i(x)\geq 0], \forall i\in \intSet{N}\\
		\sum_{i=1}^{N}&\beta_i \geq 1
		\end{split}
	\end{align}
\end{property}

\subsection{Chebychev center}
The \ac{CC} of a polyhedron $P=\{x\mid Ax\leq b\}$, with $A\in\R^{m\times n}$ and $b\in\R^m$, is the center~$x^\star\in\R^n$ of the largest ball $B(x^\star, r^\star)=\{x\mid \norm{x^\star -x} \leq r^\star\}$ contained in $P$~\cite{Boyd2004}. The radius~$r^\star$ is called the Chebychev radius.
With $A_i$ and $b_i$ being the $i$-th row of $A$ and $b$, respectively, the \ac{CC} and Chebychev radius can be computed by solving the linear program
\begin{mini!}
	{r, x }{ -r }{\label{eq:CC_base}}{}
	\addConstraint{A_i x + r\|A_i\|_2 }{\leq b_i\quad}{ i \in \intSet{m}\label{eq:CC_base_ineq}}
	\addConstraint{r}{\geq 0 }{\label{eq:CC_base_rad}}
\end{mini!}

\section{General Lane Changing Problem}
\label{sec:general_problem}
The general problem for lane changing is stated as an \ac{OCP}, similar to~\cite{Zhou2023}.
For a multi-lane environment, a total of~$\dimLanes{}$ lanes are defined by curvilinear center curves and a lane width~$\laneWidth$.
Moreover, we assume the existence of a parametric function $\gamma:\R^+ \rightarrow \R^2$ for the right-most reference lane that maps a longitudinal path coordinate~$s$ to a Cartesian point. The reference lane is parameterized by a vector~$\theta_{\gamma}$ which is included in the road geometry parameters~$\theta:=\big(\theta_\gamma,\laneWidth\big)$.
We consider a vehicle model in the Frenet coordinate frame~\cite{Ziegler2014a, Werling2010a, Reiter2023a} with states~$x(t)\in\R^{n_x}$ and inputs~$u(t)\in\R^{n_u}$, whose trajectories are governed by the nonlinear \ac{ODE}~$\dot{x}=\xi(x(t),u(t))$ with the initial condition~$x(t_0)=x_0$. Using Frenet coordinates poses mild assumptions on the maximum value of the curvature, cf.~\cite{Eilbrecht2020}.
Among others, the state of the Frenet model includes a longitudinal position state~$s$, a lateral position state~$n$, a velocity~$v$ and a heading angle mismatch~$\alpha$~\cite{Reiter2023a}.

States and controls are constrained by physical limitations depending on~$\theta$, which are expressed by admissible sets~$\stateAdmiss{}(t;\theta)$ and~$\conrlAdmiss{}(t)$.

For~$M$ vehicles on each lane, we consider $N_\mathrm{veh}=LM$~\opps{} with states $x^{\mathrm{sv}}_i(x(t),t)$ for $i\in\intSet[1]{N_\mathrm{veh}}$ that depend on the planned ego trajectory~$x(t)$. 
Note that the dependency of the states $x^{\mathrm{sv}}_i(x(t),t)$ on the ego state~$x(t)$ is due to the interaction of the ego vehicle with \opps{} and is a major source of complexity~\cite{Wang2020, Cleac2022}. 
We assume that the obstacle-free set can be approximated by~$\stateFree{}(x(t),t)$.
In the latter sections~\ref{sec:short_planner} and~\ref{sec:long_planner}, we explain how to define the set~$\stateFree{}(x(t),t)$ in an \ac{MIQP} model. 

In compliance with~\cite{Deolasee23}, multiple general objectives are proposed in the Frenet coordinate frame in order to define the desired behavior.
A goal lane index~$\refer{l}\in\intSet[1]{L}$ and a reference velocity~$\refer{v}\in\R^+$, define the goal parameters 
\begin{align*}
\Theta:=(\refer{l},\refer{v}).
\end{align*}
Curvilinear reference paths are expressed as constant lateral references~$\tilde{n}_i$ for $i\in\intSet[1]{L}$.
One desired behavior is to track the lateral lane reference the vehicle is currently driving on.
The  current reference lane index~$l(n)$ w.r.t. the current lateral state~$n$ is uniquely determined by 
\begin{equation}
	\label{eq:get_lane}
	l(n) = \bigg\lceil \frac{n}{d_\mathrm{lane}} + \frac{1}{2} \bigg\rceil.
\end{equation}
Note that determining the lane as in~\eqref{eq:get_lane} within an optimization problem is not trivial and requires for instance the use of additional integer variables, as shown in Sect~\ref{sec:short_planner}.
By using the weights~$w_n$ and~$w_v$, the cost of tracking the reference lane index~$l(n(t))$ and longitudinal reference speed~$\tilde{v}$ is 
\begin{align}
	\label{eq:cost_ocp_ref}
	\begin{split}
	g_\mathrm{ref}&(x(t),u(t); \theta, \Theta)=\\
	&w_n\Big( n(t)-\big(l(n(t))-1\big)\laneWidth\Big)^2+\\
	&w_v\Big( v(t)-\tilde{v}\Big)^2 +
	u^\top(t) R u(t),
\end{split}
\end{align}
which includes a quadratic penalty on the input~$u$, with the positive definite weighing matrix~$R\in \R^{n_u\times n_u}$.

Next, a cost for the distance to the goal lane~$\refer{l}\in\intSet[1]{L}$ with a weight~$w_\mathrm{g}\in \R^+$ is added, which is the main objective of the presented planner and written as 
\begin{align}
	\label{eq:cost_lc}
	g_\mathrm{lane}(x(t);\Theta)=w_\mathrm{g}\big|	l(n(t)) -\refer{l}\big|.
\end{align}
In the proposed approach, only lane changes towards the goal lane~$\refer{l}$ are considered.

Finally, the objective functional is
\begin{align}
	\label{eq:ocp_cost}
	\begin{split}
	&J\big(x(\cdot), u(\cdot);\theta, \Theta \big) := \\
	& \int_{t=t_0}^\infty \Big( g_\mathrm{ref}(x(t),u(t); \theta)+
	g_\mathrm{lane}(x(t);\Theta)\Big) dt,
	\end{split}
\end{align}

and the considered general optimal control problem that is approximately solved by the proposed approach is

\begin{mini!}[1]
	{x(\cdot),u(\cdot)}			
	{J\big(x(\cdot), u(\cdot);\theta,\Theta\big)}
	{\label{eq:ocp}} 
	{} % result of optimization, e.g., J^* =
	%\addConstraint{LHS.1}{RHS.1\label{Const1}}{extraConst1}
	% \breakObjective{ + \norm{x^\mathrm{F}_N-x^\mathrm{F}_{\mathrm{ref},N}}_{Q_N}^2 }
	\addConstraint{x(t_0)}{= x_0}{}
	\addConstraint{\dot{x}}{= \xi(x(t),u(t)),\quad}{t\in[t_0,\infty)}
	\addConstraint{x(t)}{\in \stateFree{}(x(t),t;\theta)\cap\stateAdmiss{}(t;\theta),\quad}{t\in[t_0,\infty)}
	\addConstraint{u(t)}{\in \conrlAdmiss{}(t),\quad}{t\in[t_0,\infty)}.
\end{mini!}
\subsection{Assumptions and simplifications}
Several assumptions and simplifications are made for the proposed planning approach in order to approximate~\eqref{eq:ocp} by an \ac{MIQP}.
As a major simplification, the vehicle dynamics are formulated by a point-mass model with mass~$m$ in a Frenet coordinate frame~\cite{Eilbrecht2020}, including the longitudinal and lateral position states~$s$ and~$n$, as well as associated velocities~$v_s$ and~$v_n$, with~$x=[s,n,v_s,v_n]^\top$, and acceleration inputs~$a_s$ and~$a_n$, with~$u=[a_s, a_n]^\top$. 
The dynamics are modeled by
\begin{equation}
	\label{eq:model}
	\dot{x}=[v_s, v_n, \frac{1}{m} a_s,\frac{1}{m} a_n]^\top.
\end{equation}
We assume the absolute value of the curvature~$\kappa(s)$ and its derivative~$\kappa^\prime(s)$ to be small for highway roads and, therefore, the acceleration in the Frenet coordinate frame is approximately equal to the acceleration in Cartesian coordinate frame~\cite{Eilbrecht2020}.
This model was empirically shown to be valid for the cases where vehicles are not driving at their dynamical limits~\cite{Miller2018} and motivated in several other works, e.g.,~\cite{Quirynen2023,Laurense2022,Burger2018,Hess2018,Schurmann2017}.
Critical evasion maneuvers are passed to a \ac{NMPC} within the presented structure.

\begin{Assumption}
	\label{as:lanechanges}
	Lane-changes of \opps{} can be detected.
\end{Assumption}
Ass.~\ref{as:lanechanges} can be satisfied by perception techniques described in~\cite{Guanetti2018, Paden2016} or by vehicle-to-vehicle communication.\\

\begin{Assumption}
	\label{as:rule_lead_follow}
	Considering two vehicles in the same lane, the rear vehicle is responsible for avoiding collisions. The leading vehicle must maintain general deceleration limits. Vehicles that change lanes must give way to vehicles on the lane they are changing to.
\end{Assumption}
Taking into account interactions among traffic participants within~$\stateFree{}(x(t),t;\theta)$ is essential for certain maneuvers in order to avoid prohibitive conservatism~\cite{Trautman2010}. 
However, the interdependence of plans among interactive agents leads to computationally demanding game-theoretic problems~\cite{Cleac2022,Buckman2019}. Similar to~\cite{Miller2018}, the leader-follower interaction is simplified by ignoring collision constraints of followers on the same lane at the current state, leaving the responsibility for collision avoidance to the follower. Other \opps{} that are not following on the current lane are considered obstacles independent of the ego plan as long as these \opps{} are on adjacent lanes or in front of the ego vehicle.
\begin{figure*}
	\begin{center}
		\includegraphics[scale=1]{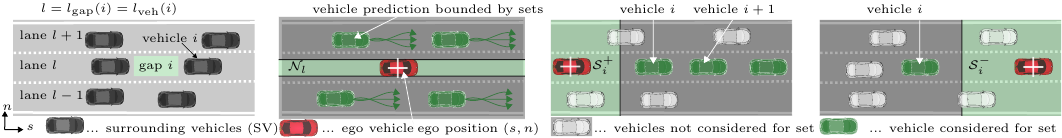}
		\caption{The first figure shows the enumeration of lanes and gaps and the rightmost three figures show the sets related to free spaces. \Acfp{SV} are uniquely enumerated. Gaps are the free spaces on a lane w.r.t. \acp{SV} and are enumerated according to the leading vehicles. An additional index is used for each frontmost gap. The sets~$\mathcal{N}_l,\mathcal{S}^+_i$ and $\mathcal{S}^-_i$ define half-spaces in the \slt{} and are plotted in \emph{green} for the position dimensions. The sets are \emph{tightened} to include all configurations of the \acp{SV} and ego vehicle to allow collision-free planning with a point-mass model. 
		All leading vehicles on the same lane are considered to construct the set $\mathcal{S}^+_{i}$, since any slower vehicle requires all following vehicles to brake.
		For the following vehicle set~$\mathcal{S}^-_{i}$ only the closest vehicle to gap~$i$ is considered, since preceding ones are assumed to not influence leading vehicles.}
		\label{fig:sketches_setting}
	\end{center}
\end{figure*}

The following assumptions consider constraints in the three-dimensional \slt{}~\cite{Deolasee23}, i.e., the space of the longitudinal and the lateral Frenet position states~$s$ and~$n$ and time~$t$.

On each lane, a maximum of~$M$ vehicles is considered. We use indices~$i\in \intSet[1]{LM}$ for the enumeration of the resulting maximum $LM$~vehicles on the lanes in ascending order, starting from lane~$l=1$ from rear to front. The free space on the back of each vehicle along the lane is referred to as \emph{gap} and enumerated according to the leading vehicle index, cf., Fig.~\ref{fig:sketches_setting}. A number of~$L$ indices are added for the gaps in front of the first vehicle on each of the $L$ lanes. Therefore the number of gap indices is~$L(M+1)$. The function~$l_\mathrm{veh}(i)$ returns the lane index of vehicle~$i$. The function~$l_\mathrm{gap}(i)$ returns the lane index of gap~$i$. The function~$M_\mathrm{lane}(i)$ returns the total number of vehicles on lane~$l=l_\mathrm{veh}(i)$ for a vehicle with index~$i$.

In the following, we assume that~$\stateFree{}(x(t),t)$ can be partitioned into sets~$\stateFree^{\mathcal{I}}(x(t),t)$ related to \opps{}~$\hat{x}_{i}^\mathrm{sv}$ within indices in the set~${i\in\mathcal{I}\subseteq \intSet[1]{LM}}$, with $\stateFree{}(x(t),t)\subseteq \stateFree^{\mathcal{I}}(x(t),t)$ and~$\stateFree(x(t),t)=\stateFree^{\intSet[1]{LM}}(x(t),t)$.

By inflating the obstacle shapes and lane boundaries by a safe distance according to all allowed configurations of the ego and the \opps{}, the planning problem can be formulated by a point-wise set exclusion of the curvilinear ego position states~$s(t)$ and~$n(t)$~\cite{LaValle2006}.

\begin{Assumption}
	\label{as:obstacle_lat}
	For all \opps{} not driving at the current ego lane~$l$, defined by the index set~$\mathcal{I}(l)=\{k\in\intSet[1]{LM} \mid l\neq l_\mathrm{veh}(k)\}$, an obstacle-free set~$\mathcal{N}_{l}=\{n\in\R\mid \lb{n}_l \leq n\leq \ub{n}_l\}$ w.r.t. the ego lateral state~$n$ can be found such that 
	\begin{equation*}
		n\in\mathcal{N}_{l} \implies x(t)\in\stateFree^{\mathcal{I}(l)}(x(t),t).
	\end{equation*}
\end{Assumption}
Ass.~\eqref{as:obstacle_lat} is used to define collision avoidance constraints to vehicles on adjacent lanes by formulating constraints on the lateral state~$n$. Without further details, it is assumed that the bounds in Ass.~\ref{as:obstacle_lat} are \emph{tight} enough to contain \emph{most} of the adjacent lanes as free space, i.e., $\mathcal{N}_{l}\neq \emptyset$.
\begin{Assumption}
	\label{as:obstacle_long_min}
	Given an \opp{} with index~$i$, upper position bounds~$\ub{s}^{\mathrm{sv}}_{i}$ and velocity bounds~$\ub{v}^{\mathrm{sv}}_{i}$ can be found that define the collision-free set
	\begin{equation}
		\label{eq:free_2d_rear}
		\mathcal{S}_{i}^-=\big\{(s,t)\in (\R^+\times\R^+)\bigm\vert s\geq\ub{s}^{\mathrm{sv}}_{i}+t\ub{v}^{\mathrm{sv}}_{i}\big\}.
	\end{equation}
	such that it holds that
	\begin{equation*}
		(t,s(t))\in\mathcal{S}_{i}^-\implies x(t)\in\stateFree^{\{i\}}(x(t),t).
	\end{equation*}
\end{Assumption}

\begin{Assumption}
	\label{as:obstacle_long_max}
	For all \opps{} on lane~$l$, with indices~$i\in\intSet[1]{M}$, lower position bounds~$\lb{s}^{\mathrm{sv}}_{i}$, velocity bounds~$\lb{v}^{\mathrm{sv}}_{i}$ and leading vehicle distances~$\Delta s_{i}$ can be found that define the set
	\begin{align}
		\label{eq:free_2d_front}
		\begin{split}	
		\mathcal{S}_{i}^+=&\bigg\{(s,t)\in (\R^+\times\R^+)\bigg| \\
		&s\leq \lb{s}^{\mathrm{sv}}_{i}+t\lb{v}^{\mathrm{sv}}_{i}+ \sum_{k=i+1}^{M_\mathrm{lane}(i)} \lb{s}^{\mathrm{sv}}_{k}+t\lb{v}^{\mathrm{sv}}_{k}-\Delta s_{k-1}\bigg\}.
		\end{split}
	\end{align}
	such that for $0\leq t\leq \ub{t}$ it holds that
	\begin{equation*}
		(t,s(t))\in\mathcal{S}_{i}^+\implies x(t)\in\stateFree^{\{i,\ldots,M\}}(x(t),t).
	\end{equation*}
\end{Assumption}

Similar assumptions are made in related work, e.g.,~\cite{Johnson2013, Miller2018}, and with a more accurate lateral shape in~\cite{Quirynen2023}. 
\begin{figure}
	\begin{center}
		\includegraphics[scale=1]{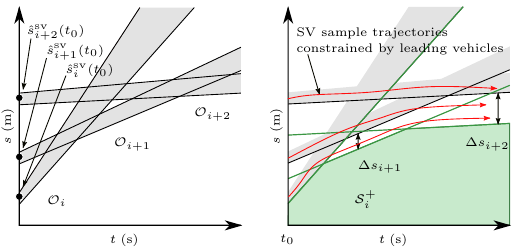}
		\caption{Construction of longitudinal ostacle-free space~$\mathcal{S}_i^+$ for an \opp{} with index~$i$ and two leading vehicles. The left plot shows the nominal prediction sets~$\mathcal{O}_i$. The right plot shows the blocking lower-bound set, enforced on the following vehicles. Red trajectories are plotted corresponding to samples of actually driven trajectories.}
		\label{fig:sketches_region}
	\end{center}
\end{figure}
The bounds in Ass.~\ref{as:obstacle_long_max} approximate a distribution that is generated by the \emph{intelligent driver model}~\cite{Treiber2000}, where a vehicle either drives or approaches a range around a reference velocity~$\refer{v}^\mathrm{sv}_i$, with $\lb{v}^\mathrm{sv}_i\leq\refer{v}^\mathrm{sv}_i\leq\ub{v}^\mathrm{sv}_i$, or drives within a certain distance~$\Delta s_{i}$ to a slower leading vehicle~\cite{Johnson2013, Miller2018}, cf., Fig.~\ref{fig:sketches_region}.
The set 
\begin{equation*}
	\mathcal{O}_i=\big\{(s,t)\in (\R^+\times\R^+)\bigm\vert \lb{s}^{\mathrm{sv}}_{i}+t\lb{v}^{\mathrm{sv}}_{i} \leq s\leq \ub{s}^{\mathrm{sv}}_{i}+t\ub{v}^{\mathrm{sv}}_{i}\big\}
\end{equation*}
is referred to the nominal \opp{} prediction set in the absence of leading vehicles. In each planning step, the bounds of $\mathcal{O}_i$ are updated based on the current \opp{} state~$\hat{x}^\mathrm{sv}_i$, where for the velocity bounds it additionally holds that~$\lb{v}^\mathrm{sv}_i\leq\hat{v}^\mathrm{sv}_i\leq\ub{v}^\mathrm{sv}_i$ and for the position bounds $\lb{s}^\mathrm{sv}_i\leq\hat{s}^\mathrm{sv}_i\leq\ub{s}^\mathrm{sv}_i$ holds.

\begin{Assumption}
	\label{as:lane_change_time}
	The duration of a lane-change~$t_\mathrm{lc}$ is upper-bounded by~$t_\mathrm{lc}\leq\ub{t}_\mathrm{lc}$.
\end{Assumption}
For a concise notation, no offsets are assumed, i.e., the current lane and gap index are~$1$, the current planning time is assumed at zero seconds and the initial lateral reference and longitudinal estimated state are set to~$0$, therefore~$\tilde{n}_0=0$ and~$\hat{s}=0$. 

\subsection{Obstacle-free set approximations}
In the following, convex obstacle-free sets in the \slt{} are defined as intersections of the sets $\mathcal{S}_{i}^+,\mathcal{S}_{i}^-$ and $\mathcal{N}_{l}$, which serve as a basis for the proposed \lstp{}.

\begin{figure}
	\begin{center}
		\includegraphics[scale=1]{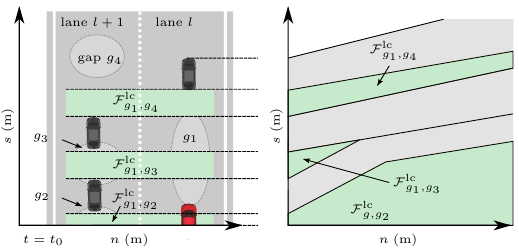}
		\caption{Sketch of obstacle-free sets~$\freeSpace^\mathrm{lc}$ (green) for lane changing related to three \opps{} on the two lanes~$l$ and~$l+1$. The left plot shows the curvilinear space with coordinates~$s$ and~$n$. The right plot shows the \st{}. Three possible gaps with indices~$g_2$, $g_3$, and $g_4$ on the consecutive lane~$l+1$ are available for a transition from gap index~$g_1$ and lane~$l$. }
		\label{fig:sketches_lc_sets}
	\end{center}
\end{figure}

Two convex three-dimensional sets in the \slt{} are used to formulate a lane change from lane~$l$ and gap index~$g$ on the same lane, i.e., $l=l_\mathrm{gap}(g)$, to gap index~$g^+$ on the next lane~$l+1$, i.e., $l+1=l_\mathrm{gap}(g^+)$.
First, for lane-keeping, obstacle avoidance reduces to the problem of staying within the current lane boundaries (Ass.~\ref{as:obstacle_lat}), ignoring following \opps{} on the same lane (Ass.~\ref{as:rule_lead_follow}) and consider leading \opps{}, with an upper-bound related to~\eqref{eq:free_2d_front}, stated as the convex obstacle-free set over the longitudinal and lateral position and time 
\begin{equation}
	\label{eq:free_3d_front}
	\freeSpace^{+}_{g}= 
	\left\{
	(t,s,n) \mid(t,s) \in \mathcal{S}^+_{g}, \quad n \in \latspace{l_\mathrm{gap}(g)}
	\right\}.
\end{equation}
Next, the free set for a lane change is defined in the two-dimensional \st{}, which is a subspace of the \slt{}, as
\begin{align}
	\begin{split}
		\label{eq:free_2d_lc}
		\freeSpaceTwoDim^\mathrm{lc}_{g,g^+}&= 
		\big\{
		(t,s) \in \mathcal{S}^+_{g} \cap \mathcal{S}^+_{g^+} \cap \mathcal{S}^-_{g^+}
		\big\}.
	\end{split}
\end{align}
Finally, as shown in Fig.~\ref{fig:sketches_lc_sets}, the free space related to a lane change from lane~$l$ and the related gap index~$g$ to lane~$l+1$ and the related gap index~$g^+$ is
\begin{align}
	\label{eq:free_3d_lc}
	\begin{split}
	\freeSpace^{lc}_{g,g^+}= \big\{&
	(t,s,n) \in \freeSpaceTwoDim^\mathrm{lc}_{g,g^+} \times \R \; \big\vert \\ & 
	n \in\mathrm{conv}( \latspace{l_\mathrm{gap}(g)} \cup \latspace{l_\mathrm{gap}(g^+)})\big\}.
	\end{split}
\end{align}
For a lane change, both lanes are required to be free of \opps{} and for the next lane~$l+1$, also rear vehicles need to be considered for the duration of the lane change, cf. Ass.~\ref{as:rule_lead_follow}.
Only the closest rear vehicle on the next lane needs to be considered, since more distant vehicles are constrained by preceding ones, cf. Fig.~\ref{fig:sketches_lc_sets}.

The convexity of~\eqref{eq:free_3d_front}, \eqref{eq:free_2d_lc} and \eqref{eq:free_3d_lc} stems from the fact that each set is an intersection of hyperplanes, which implies convexity~\cite{Boyd2004}.
In case of a detected lane change of an obstacle, which we assume to be detectable (Ass.~\ref{as:lanechanges}), both lanes are considered to be blocked for the whole prediction horizon, cf. Ass.~\ref{as:lanechanges}.

\section{Short-Horizon Approximations}
\label{sec:short_planner}
The \acf{STF} approximates the vehicle dynamics for a prediction horizon~$\dimT$ and a maximum of one lane change towards the goal lane~$\refer{l}$, similar to~\cite{Miller2018,Quirynen2023}. The selection of the particular gap index~$g^+$ on the next lane is part of the \acf{LTF}, which is vice versa constrained by the trajectory of the \shortHorPlanner{} in the final \lstp{} formulation.

	Discretizing the model~\eqref{eq:model} with a discretization time~$t_\mathrm{d}$ yields the linear discrete-time model
	\begin{equation}
		\label{eq:dynamics}
		x_{k+1}=Ax_k+Bu_k.
	\end{equation}
	A prediction horizon~$\tHor{}=\nHor{} t_\mathrm{d}$ with $\nHor$ steps is used to approximate the infinite horizon in \eqref{eq:ocp}.

We define the acceleration bounds on the Frenet coordinate frame accelerations by the admissible control set 
\begin{equation}
	\label{eq:control_set}
	\conrlAdmiss{}=\big\{u\in\R^{n_u} \bigm\vert \lb{u} \leq u_k \leq \ub{u} \big\},
\end{equation}
where $\lb{u}=[\lb{a}_\mathrm{lon},-\ub{a}_\mathrm{lat}]^\top$ and  $\ub{u}=[\ub{a}_\mathrm{lon},\ub{a}_\mathrm{lat}]^\top$. Nonetheless, higher curvatures and its derivatives can be inner-approximated by convex sets according to~\cite{Eilbrecht2020}.
Moreover, the constraint
\begin{equation}
	\label{eq:state_set}
	\alpha_{\mathrm{l}}v_{s}\leq v_{n}\leq\alpha_{\mathrm{r}}v_{s}
\end{equation}
limits the lateral velocity in order to approximate the nonholonomic motion of a kinematic vehicle model.

The reference tracking cost~\eqref{eq:cost_ocp_ref} is approximated for the \shortHorPlanner{}, whereas the remaining costs of the objective~\eqref{eq:ocp_cost} are approximated as part of the \longHorPlanner{}.
Binary variables~$\lambda_k\in \intSet[0]{1}$, with $\Lambda=[\lambda_0,\ldots,\lambda_{N}]$, are used to indicate whether the planned position is on the current lane, $\lambda_k=0$, or on the next lane, $\lambda_k=1$. The lane indices are always updated w.r.t. the current state such that $\lambda_k=0$ corresponds to the current lane. 
A lateral reference can therefore be expressed by
\begin{subequations}
	\label{eq:lane_change}
\begin{align}
	\tilde{n}_{k}&=\laneWidth\lambda_k, \ k\in \intSet[0]{N},\\
	\lambda_{k+1}&\geq\lambda_k,\ k\in \intSet[0]{N-1}.
	\label{eq:lane_change_helper}
\end{align}
\end{subequations}
Constraint~\eqref{eq:lane_change_helper} is used to cut off binary assignments to ease the solution of the \ac{MIQP} problem. The constraint
\begin{equation*}
	\tilde{n}_{k}-\frac{\laneWidth}{2} \leq n_k \leq \tilde{n}_{k}+\frac{\laneWidth}{2}
\end{equation*}
is added to guarantee that from $\tilde{n}_{k}> 0$ the planned ego vehicle state is located on the next lane.
Cost~\eqref{eq:cost_ocp_ref} can consequently be approximated with Frenet states as
\begin{align}
	\label{eq:cost_sh_ref}
	\begin{split}
	g_\mathrm{ref}^\mathrm{st}&(x_k,u_k,\lambda_k)=\\
	&w_n (\laneWidth\lambda_k-n_k)^2 +
	w_v  (\tilde{v}-v_{\mathrm{s},k})^2+
	u^\top_k R u_k.
\end{split}
\end{align}
\begin{figure*}
	\begin{center}
		\includegraphics[trim={7mm 5mm 0mm 5mm},clip,width=55mm]{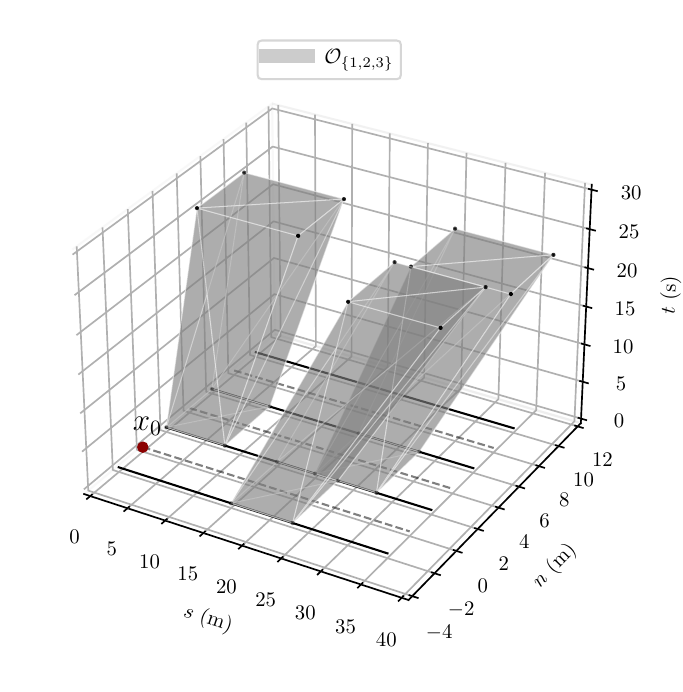}
		\includegraphics[trim={7mm 5mm 0mm 5mm},clip,width=55mm]{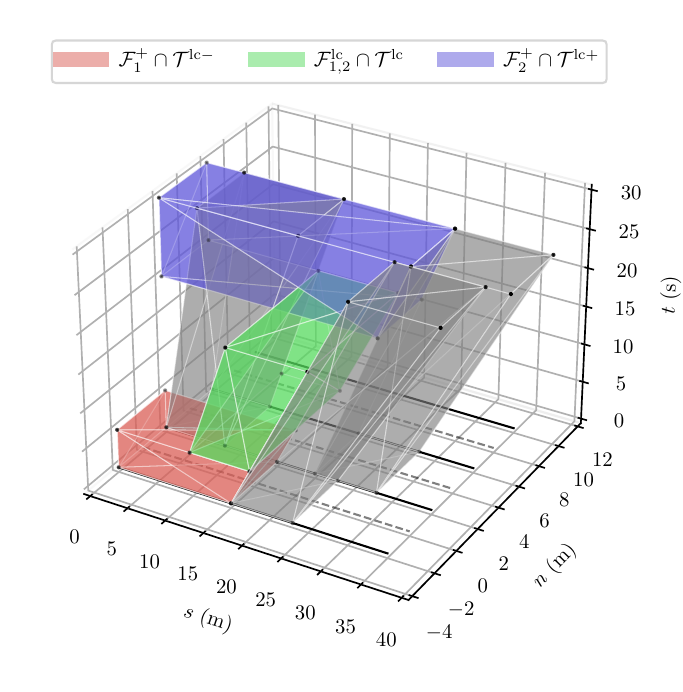}
		\includegraphics[trim={7mm 5mm 0mm 15mm},clip,width=55mm]{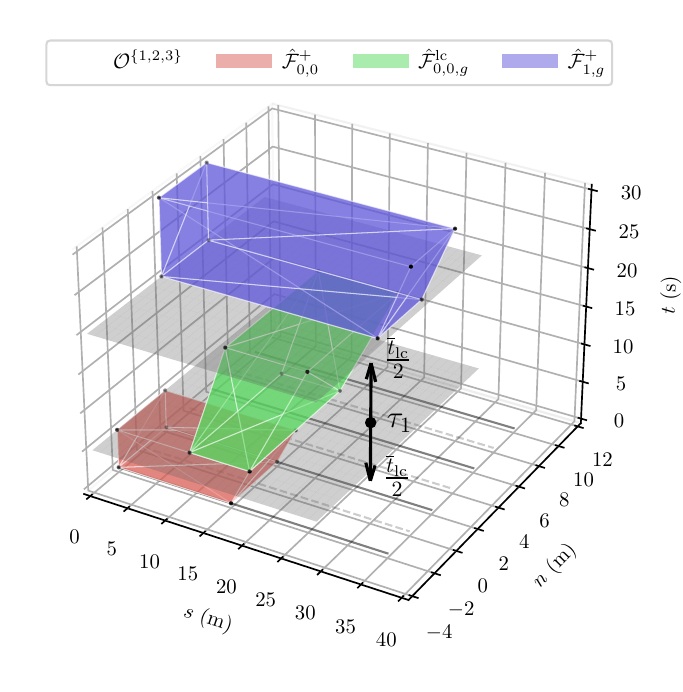}
		\caption{Visualization of \acp{SV} sets~$\occspace_{\{1,2,3\}}$ in the \slt{} and consecutive convex free regions between gap index~$1$ and gap index~$2$. The time sub-spaces~$\mathcal{T}^\mathrm{lc-}=\{(t,s,n)|t\leq \tau_1-\ub{t}_\mathrm{lc}/2\}$, $\mathcal{T}^\mathrm{lc}=\{(t,s,n)|\tau_1-\ub{t}_\mathrm{lc}/2 \leq t\leq \tau_1+\ub{t}_\mathrm{lc}/2\}$ and $\mathcal{T}^\mathrm{lc+}=\{(t,s,n)|t\leq\tau_1+\ub{t}_\mathrm{lc}/2\}$ define the consecutive time-related spaces on the planning horizon. The set $\freeSpace^{+}_{1}\cap\mathcal{T}^\mathrm{lc-}$ is the obstacle-free space on the first lane before the lane change, $\freeSpace^\mathrm{lc}_{1,2}\cap\mathcal{T}^\mathrm{lc}$ is the free space during the lane change and $\freeSpace^{+}_{2}\cap\mathcal{T}^\mathrm{lc+}$ is the obstacle-free space on the next lane after the lane change. Rear vehicles in the same lane are ignored, i.e., a vehicle is always allowed to brake. The binary variables~$\lambda_k$ determine, which set constraints are active for each~$x_k$.}
		\label{fig:polyhedron}
	\end{center}
\end{figure*}

Cost~\eqref{eq:cost_sh_ref} includes the term~$(\laneWidth\lambda_k-n_k)^2=(\laneWidth\lambda_k)^2 -2\laneWidth\lambda_kn_k + n_k^2 $ with the bilinear term~$\minus 2\laneWidth\lambda_k n_k$ which cannot directly be handled by \ac{MIQP} solvers~\cite{Gurobi2023}. Thus, this bilinear term is reformulated by introducing continuous auxiliary variables~$\slack_{\mathrm{bin},k}\in\R^+$, the additional constraint~$\slack_{\mathrm{bin},k}=\lambda_k n_k$ and further related constraints according to Property~\ref{def:bilinear}.\\

Finally, safety constraints approximating the set~$\stateFree(x(t),t)$ for the current and the next lane are formulated by considering~$M$ vehicles on the current lane, which is always set to~$l=1$ and the next lane~$l^+=2$ and a chosen gap index~$g^+\in\Z$ on the next lane, with~$2=l_\mathrm{gap}(g^+)$. 

Changing lane at time~$\tau_1$ follows three stages (cf. also~\cite{Miller2018}) where, in each stage, the constraints can be formulated as convex sets, cf., Fig.~\ref{fig:polyhedron}.
First, at time~$t\leq \tau_1 - \frac{1}{2}\ub{t}_\mathrm{lc}$, the ego lane is tracked, second the lane is changed until the time limit~$\tau_1 + \frac{1}{2}\ub{t}_\mathrm{lc}$, and thirdly constraints for driving on the next lane hold for $t\geq \tau_1 + \frac{1}{2}\ub{t}_\mathrm{lc}$.
The lane change time on either lane is approximated by the upper-bound related to the time indices, $n_\mathrm{lc}=\lceil\frac{\ub{t}_\mathrm{lc}}{2 t_\mathrm{d}}\rceil$.
Consequently, the lane change phases can be formulated in terms of index shifts of~$\lambda_k$, with~$[(1-\lambda_{k+n_{lc}})= 1]$ indicating the first stage, $[(\lambda_{k+n_{lc}}-\lambda_{k-n_{lc}})= 1]$ indicating the transition phase and $[ \lambda_{k-n_{lc}}= 1]$ indicating the last stage on the next lane, cf. Fig.~\ref{fig:polyhedron}.
For out-of-range indices, i.e., $k<0$ or $k>N$, the first value~$\lambda_0$ or the last value~$\lambda_N$ are padded.
For each position and time tuples~$(t_k, s_k, n_k)$ with $k\in \intSet[0]{N}$ and the current gap index~$g=1$, we require
\begin{subequations}
\label{eq:lane_change_bin}
\begin{align}
	[1-\lambda_{k+n_{lc}}=1]& \implies (t_k, s_k, n_k) \in \freeSpace^{+}_{1},\\
	[\lambda_{k+n_{lc}}-\lambda_{k-n_{lc}}=1]& \implies (t_k, s_k, n_k) \in \freeSpace^\mathrm{lc}_{1,g^+},\\
	[\lambda_{k-n_{lc}}=1]& \implies (t_k, s_k, n_k) \in \freeSpace^{+}_{g^+},
\end{align}
\end{subequations}
where the implications are reformulated according to Property~\ref{def:bigm_activation}.

Recursive feasibility requires disjunctive terminal velocity constraints depending on the final lane, which is either the current lane implying~$[ \lambda_N=0]$ or the next lane, implying $[ \lambda_N=1]$. Therefore, the terminal set is expressed by
\begin{subequations}
\label{eq:lane_change_terminal}
\begin{align}
	[\lambda_N=1]&\implies  v_{s,N}\leq \lb{s}^\mathrm{sv}_{g^+} + t_\mathrm{d}N \lb{v}^\mathrm{sv}_{g^+},\\
	[1-\lambda_N=1]&\implies v_{s,N}\leq \lb{s}^\mathrm{sv}_{1} + t_\mathrm{d}N \lb{v}^\mathrm{sv}_{1},\\
	\quad v_{n,N} &= 0.
\end{align}
Note that this terminal set is restrictive since it upper-bounds the final velocity with the velocity of the preceding vehicle on the respective lane. An increased terminal safe set could be formulated by piece-wise linear approximations of deceleration constraints which requires further binary variables, cf.~\cite{Miller2018}.\\

So far, constraints and costs have been introduced that are used as part of the \shortHorPlanner{} to plan a collision-free discrete-time trajectory from the current lane to a certain gap at the next lane. Noteworthy, this trajectory is constrained such that it is always safe w.r.t. the obstacle constraints. The selection of the possible gap indices and also all further gaps towards the goal lane are formulated in the \longHorPlanner{} and explained in the next section. The \shortHorPlanner{} and the \longHorPlanner{} are formulated in the final \ac{MIQP} with mutual constraints, such that the rather approximate \longHorPlanner{} cannot plan transitions that are infeasible w.r.t. the \shortHorPlanner{}.
\end{subequations}
\section{Long-Horizon Approximations}
\label{sec:long_planner}
Within the~\longHorPlanner{}, costs and constraints are formulated that select collision-free transition gaps between two adjacent lanes.
For long horizons, a fixed discretization in time is prohibitive, since the number of variables increases with the horizon length and would make the optimization problem hard to solve~\cite{Quirynen2023}.
To circumvent the computational scaling with the prediction time, we propose a formulation in the two-dimensional continuous \st{}, where we exclusively model the transitions as points in time and longitudinal position for each lane change, with the transition times~$T=[\tau_1,\ldots,\tau_{L-1}]^\top$ and 
longitudinal transition positions~$\Sigma=[\sigma_1\ldots,\sigma_{L-1}]^\top$.

In the following, three synergetic concepts are formulated to approximate the transitions, i.e., constraints that approximate reachability, a formulation for guaranteeing and maximizing the distance to \opps{} and a disjunctive formulation for choosing among gaps between vehicles for each lane. Binary variables are used to indicate whether transitions are invalid, resulting in the tuples of valid transitions for~$\bar{L}\leq L$ lanes.

\subsection{Approximate reachability}
\label{sec:planner_long_reachability}
Reachability between transitions is approximated by the set~$\reachspace{}(\tau_l,\sigma_l)$ using constraints defined by operating velocity bounds~$\lb{v}_\mathrm{op}$ and $\ub{v}_\mathrm{op}$ and an approximation~$\refer{t}_\mathrm{lc}$ of the time required to traverse a lane. 
The operating velocity bounds are artificially added to approximate the \emph{true} reachable set around the expected velocity range of the vehicle.

\begin{figure}
	\begin{center}
		\includegraphics[scale=1]{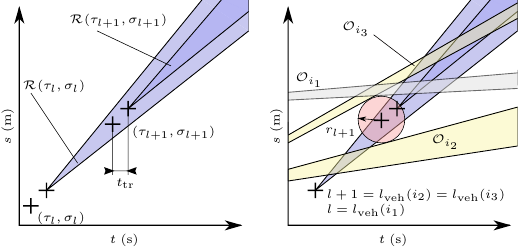}
		\caption{The left plot shows the reachable sets after the transition to lane~$l$ and after the transition to lane~$l+1$. The reachable set has an offset related to the estimated traversal time~$\refer{t}_\mathrm{lc}$. The right plot shows the \acf{CC} of the transition to lane~$l+1$ from gap index~$i_1$ to gap index~$i_3$ with two \acp{SV} on the next lane~$l+1$ (yellow) and one \ac{SV} on the current lane~$l$ (grey). The time axis is scaled by the reference velocity~$\refer{v}$, which is here assumed to be~$1$.}
		\label{fig:sketches_reach}
	\end{center}
\end{figure}

The approximated set is used to define constraints for the next transition~$(\tau_{l+1},\sigma_{l+1})$, cf. Fig.~\ref{fig:sketches_reach}. Each reachable set depends on the last transition~$(\sigma_l,\tau_l)$ by the shifted cone
\begin{equation}
	\label{eq:reach}
	\reachspace{}(\tau_l,\sigma_l)  = 
	\left\{
	(\tau_{l+1},\sigma_{l+1})\middle\vert
	\begin{array}{@{}l@{}}
		\sigma_{l+1} \leq \sigma_{l} +\ub{v}_\mathrm{op} (\tau_{l+1}-\tau_l-\refer{t}_\mathrm{lc}) \\
		\sigma_{l+1} \geq \sigma_{l} +\lb{v}_\mathrm{op} (\tau_{l+1}-\tau_l+\refer{t}_\mathrm{lc})
	\end{array}
	\right\}.
\end{equation}
The convex reachable set~\eqref{eq:reach} is an approximation using the velocity bounds~$\ub{v}_\mathrm{op}$ and $\lb{v}_\mathrm{op}$. Using bounds on the acceleration would result in nonconvex quadratic constraints which could be still approximated by the problem specific parameters~$\refer{t}_\mathrm{lc}, \ub{v}_\mathrm{op}$ and $\lb{v}_\mathrm{op}$.
\subsection{Chebychev centering for transitions}
\label{sec:planner_long_chebey}
Next, criteria for determining the locations of continuous transition points are defined.
Transitions require an obstacle-free area for a minimum of the duration of the lane-change~$\ub{t}_\mathrm{lc}$, as defined for the \shortHorPlanner{}.
Beyond the minimum required time, an approach based on the \acf{CC} is proposed that centers the transition in the \st{} related to obstacle constraints, i.e., the transition should be planned at a maximum weighted distance to constraints.

The \ac{CC} formulation of~\eqref{eq:CC_base} is used as a basis for further considerations. 
Centering constrains for the polytope defined by~$[\tau,\;\sigma]\in\mathcal{S}^+_{g}$ are written as ~$h^+_{g}(\tau, \sigma, r)\leq0$ according to~\eqref{eq:CC_base}, which includes the centering radius~$r$. For the polytope defined by~$[\tau,\;\sigma]\in\mathcal{S}^-_{g} \cap\mathcal{S}^-_{g} $, the centering constraints are denoted by~$h_{g}(\tau, \sigma, r)\leq0$.

Notice, that the \st{} has different units, namely longitudinal distance and time. To achieve a meaningful distance measure, the time coordinate is scaled by the reference velocity~$\refer{v}$. Therefore the unit of the radius is Meters.

The constraint~\eqref{eq:CC_base_rad} is tightened to~$r\geq\lb{r}$ to guarantee the minimum distance to obstacle constraints in the \st{}, i.e., the centering is only feasible for a centering radius higher than a threshold $\lb{r}$. 
Using the sequence of gap indices~$[g_1, \ldots, g_{L-1}]$, with $l=l_\mathrm{gap}(g_l)$, for each lane transition and the accumulated cost
\begin{equation*}
	G_\mathrm{safe}(R) =- w_\mathrm{safe} \sum_{l=1}^{\dimLanes-1}r_l,
\end{equation*}
with transition radii~$R=[r_1,\ldots,r_{L-1}]^\top$ and a weight~$w_\mathrm{safe}$ to promote a further safety distance beyond the hard constraints, all transitions can be formulated in the shared linear program 
\begin{mini!}
	{R , \Sigma,T }{ G_\mathrm{safe}(R) }{}{\label{eq:cc_standard}}
	\addConstraint{	h^+_{g_l}(\tau_l, \sigma_l, r_l)}{\leq 0, }{\quad l \in \intSet[1]{L-2}}
	\addConstraint{	h_{g_{l}}(\tau_{l}, \sigma_{l}, r_{l})}{\leq 0, }{\quad l \in \intSet[2]{L-1}}
	\addConstraint{\lb{r}}{\leq r_l,}{\quad l \in \intSet[1]{L-1}.}
\end{mini!}
Fig.~\ref{fig:sketches_reach} shows the centering of a transition~$(\tau_{l+1}, \sigma_{l+1}, r_{l+1})$ from lane~$l$ to lane~$l+1$ in the presence of a leading \opp{}~$i_1$ on lane~$l$ with~$l=l_\mathrm{veh}(i_1)$ and two \opps{}~$i_2,i_3$ on the next lane~$l+1$ with~$l+1=l_\mathrm{veh}(i_2) =l_\mathrm{veh}(i_3)$. Besides the \opps{} constraints, the center of the transition~$(\tau_{l+1}, \sigma_{l+1})$ is constrained by the approximated reachable set.\\
The linear program~\eqref{eq:cc_standard} is not directly solved but its cost and constraints are included in the final \lstp{} \ac{MIQP}. Notice that, therefore, the centering is solved as a weighted trade-off to other constraints such as the duration of the lane change. The sequence of gap indices is determined by the disjunctive formulation including the constraints within a "big-M" formulation, cf. Property~\ref{def:disjunction}, and explained in the next Section~\ref{sec:planner_long_disjunctive}.
Notice that computing the transitions purely by maximizing the distance to \opps{}, without including a measure along the time axis, would ignore the safety distance related to the relative velocity of vehicles.

\subsection{Disjunctions among gaps}
\label{sec:planner_long_disjunctive}
A fundamental combinatorial aspect of lane change planning is the choice of gap indices on each lane. In Sect.~\ref{sec:planner_long_reachability}, it was shown how to constrain transitions to an approximate reachable set and in Sect.~\ref{sec:planner_long_chebey}, a formulation to center a transition in the \st{} was introduced, given a sequence of gap indices.
As an essential final component of the \longHorPlanner{}, a disjunctive formulation of choosing a single gap on each lane is proposed according to Property~\ref{def:disjunction}.
To activate constraints related to a certain gap, on each transition~$(\tau_{l},\sigma_{l})$ binary variables~$\beta_{i}\in\intSet[0]{1}$ are used and summarized in the vector~$B\in  (\intSet[0]{1})^{(L-1)(M+2)}$. The activation of gaps starts on the second lane since the current lane gap is trivially fixed.
In each lane, one additional binary variable is added to account for the option of \emph{no transition} or \emph{lane-keeping}. 
Therefore, this particular binary variable with index~$\hat{g}$ implies the variables~$(\tau_{l},\sigma_{l})$ to be unconstrained by defining $h_{\hat{g}}(\tau_{l}, \sigma_{l}, r_{l})\leq M$, where~$M$ is a large number. 
For the following definitions, we define the set~$\mathcal{G}_l:=\{g\mid l=l_\mathrm{gap}(g)\}$, i.e., the set of all~$gap$ indices on a particular lane, including the additional virtual unconstrained one and the set~$\hat{\mathcal{G}}\in \intSet[1]{L-1}$ that contains all indices of unconstrained added gaps.
The disjunctions 
\begin{align}
	\label{eq:disjunction_pre}
	\bigvee_{g^+\in \mathcal{G}_{l+1}} \begin{bmatrix}
		\beta_{g^+}\\
		h_{g^+}(\tau_{l}, \sigma_{l}, r_{l})\leq 0 \\
	\end{bmatrix},
	\quad \forall l \in \intSet[1]{\dimLanes-1},
\end{align}
constrain the transition~$(\tau_{l}, \sigma_{l})$ onto lane~$l+1$ by the leading and following vehicles of a selected gap index~$g^+$, where~$\beta_{g^+}=1$ and the disjunctions
\begin{align}
	\label{eq:disjunction_post}
	\bigvee_{g\in\mathcal{G}_{l}}\begin{bmatrix}
		\beta_{g}\\
		h^+_{g}(\tau_l, \sigma_l, r_l)\leq 0 
	\end{bmatrix},
	\quad \forall l \in \intSet[2]{\dimLanes-1},
\end{align}
constrains the transition~$(\tau_{l}, \sigma_l)$ from lane~$l$ to the next lane only by the leading vehicles.
In the case of \emph{no transition}, i.e., the additional binary variables~$\beta_{\hat{g}}$ is activated, where~$\hat{g}$ the index of the virtual added unconstrained gap, a high cost~$G_\mathrm{lane}(\Sigma,T,B)$ is added that approximates \eqref{eq:cost_lc} for not changing the lane.

Note that the binary variables~$B$ are related to the gaps on each lane, starting with the second lane~$l=2$. The transitions~$(\tau_{l}, \sigma_{l})$ are related to two adjacent lanes~$l$ and~$l+1$, starting with the transition from the first lane~$l=1$. This distinction is crucial to the disjunctive constraints for each transition. For a transition~$l$ related to the departing lane~$l$, only leading vehicle constraints~$h^+_{g}(\tau_l, \sigma_l, r_l)\leq 0 $ related to gap index~$g$ are considered according to Ass.~\ref{as:rule_lead_follow}. For the next lane gap index~$g^+$ both, the preceding constraints and the leading vehicle constraints in~$h_{g^+}(\tau_l, \sigma_l, r_l)\leq 0 $ are used. This formulation models interactive behavior by allowing to slow down \opps{} on the current lane to reach a certain gap on the next lane. 

The following further constraint on the binary variables
\begin{equation}
	\label{eq:disjunction_bin1}
	\sum_{g\in\mathcal{G}_{l}} \beta_{g} =1,\quad\forall l \in \intSet[2]{\dimLanes},
\end{equation}
reduces the search space for the \ac{MI} solver. 
For each pair of consecutive virtual gaps~$\hat{g}$ and~$\hat{g}^+$, with $l_\mathrm{gap}(\hat{g}^+)=l_\mathrm{gap}(\hat{g})+1$, the physical constraint
\begin{equation}
	\label{eq:disjunction_bin2}
	\beta_{\hat{g}^+}\geq\beta_{\hat{g}},
\end{equation} 
sets all further lane-changes~$l> l_1$ to \emph{lane-keeping}, if a lane is blocked.

\subsection{Cost approximations}
In the following, a lane changing cost~$G_\mathrm{lane}(\Sigma,T,B)$ that approximates~\eqref{eq:cost_lc} and a reference velocity cost~$g^\mathrm{lh}_\mathrm{ref}(\Sigma,T)$ that approximates~\eqref{eq:cost_ocp_ref} and penalizes transitions with deviations of the reference velocity~$\refer{v}$ are formulated.\\

Cost~\eqref{eq:cost_lc} linearly penalizes the number of lanes distant from the goal lane and is integrated over time in the final objective.
This integral can be approximated over a horizon~$t_f$ by the following sum
\begin{align}
	\label{eq:cost_app_lane}
	\begin{split}
	\int_{t=t_0}^{t_f} &g_\mathrm{lane}(x(t);\Theta)  dt \approx\\
	&G_\mathrm{lane}(\Sigma,T,B):= w_g \sum_{\hat{g}\in \hat{\mathcal{G}}}  \tau_l (1-\beta_{\hat{g}}) + t_f \beta_{\hat{g}},
	\end{split}
\end{align}
which penalizes the duration~$\tau_l$ on each lane~$l$, if there was a valid transition, i.e., $\beta_{\hat{g}}=0$.
If \emph{no transition} was computed for lane~$l$, i.e., $\beta_{\hat{g}}=1$, the cost for the full horizon staying on the lane is summed in \eqref{eq:cost_app_lane}.
Note that~\eqref{eq:cost_app_lane} contains bilinear terms of binary variables~$\beta_{\hat{g}}$ and~$\sigma_{l}$, both decision variables.
The terms are treated by introducing an additional variable~$\slack_{\mathrm{bi},l}$ for each bilinear term, cf., Property~\ref{def:bilinear}. 
All auxiliary variables related to bilinear terms are summarized by $\Slack_{\mathrm{bi}}=[\slack_{\mathrm{bi},1}, \ldots, \slack_{\mathrm{bi},L},\slack_{\mathrm{bin},1},\ldots, \slack_{\mathrm{bin},N-1}]$

Finally, the difference of the reference speed according to~\eqref{eq:cost_ocp_ref} is penalized for two consecutive transitions by
\begin{align}
	\label{eq:cost_app_speed}
	\begin{split}
	G_\mathrm{ref}(T,\Sigma) =& \frac{w_v t_f}{l_g}
	\Bigg(\sum_{l=2}^{l_g}\big((\sigma_{l}-\sigma_{l-1})+(\tau_l-\tau_{l-1})\refer{v}\big)^2
	\Bigg).
\end{split}
\end{align}
Cost~\eqref{eq:cost_app_speed} approximates the duration between lane transitions with the constant value~$\frac{t_f}{l_g}$, starting from the second transition as the reference cost approximation of~\eqref{eq:cost_ocp_ref} for the first transition is included in the \ac{STF} cost~\eqref{eq:cost_sh_ref}.

Notably, this cost approximation for the reference velocity neglects the planned time driving on each lane.

\section{Long-Short-Horizon Motion Planner}
\label{sec:long_short_planner}
In the following, we complete the final motion planning \ac{MIQP} problem with necessary additional formulations for combining the \shortHorPlanner{} of Sect.~\ref{sec:short_planner} and the \longHorPlanner{} of Sect.~\ref{sec:long_planner}.

First, the formulations of the \shortHorPlanner{} and the \longHorPlanner{} are combined \emph{consistently} according to the following definition for the first transition~$(\tau_1,\sigma_1)$. 
\begin{definition}
	\label{def:consistency}
	A transition~$(\tau, \sigma)$ is consistent with the longitudinal states~$s_k$ and the lateral states~$n_k$ of a trajectory, with $k\in\intSet[0]{N}$, if and only if the following inequalities hold
	\begin{subequations}
	\begin{align}
		n_k \leq \frac{\laneWidth}{2}, \forall k \in \{i \in \intSet[0]{N}\mid i \td{} \leq \tau\},\\
		n_k > \frac{\laneWidth}{2}, \forall k \in \{i \in \intSet[0]{N}\mid i \td{} > \tau\},\\
		s_k \leq \sigma, \forall k \in \{i \in \intSet[0]{N}\mid i \td{} \leq \tau\},\\
		s_k > \sigma, \forall k \in \{i \in \intSet[0]{N}\mid i \td{} > \tau\}.
	\end{align}
	\end{subequations}
\end{definition}
Def.~\ref{def:consistency} states that the position states of the \shortHorPlanner{} trajectory must be located on the current lane, closer than the longitudinal position~$\sigma$ and before the transition time~$\tau$ and on the consecutive lane and position, thereafter. 

In the following, the consistency formulation for the first transition~$(\tau_1,\sigma_1)$ of a feasible solution of the \lstp{} is shown.
Therefore, constraints among the discrete decision variables~$\lambda_k$, the transition~$(\tau_1,\sigma_1)$ and positions~$s_k$ of the \shortHorPlanner{} are defined by the pair-wise \emph{exclusive} disjunctions according to Property~\ref{def:disjunction}, 
\begin{align}
\label{eq:sh_lh_coupling_constraints}
	\begin{bmatrix}
		[\lambda_k=1] \\
		k t_d \geq \tau_1\\
		s_k \geq \sigma_1
	\end{bmatrix}
	\vee
	\begin{bmatrix}
		[\lambda_k=0] \\
		k t_d < \tau_1\\
		s_k < \sigma_k
	\end{bmatrix},
	\quad \forall k \in \intSet{N},
\end{align}
For each pair~$k$, the disjunctions~\eqref{eq:sh_lh_coupling_constraints} use the same binary variable~$\lambda_k$, yet, with the opposite indication, i.e., either $[\lambda_k=0]$ or $[\lambda_k=1]$, which makes it exclusively choosing the related constraints.

Moreover, a terminal set formulation for the \shortHorPlanner{} is required to reach a transition~$(\tau_1, \sigma_1)$ with $\tau_1\geq N t_\mathrm{d}$, i.e., the transition time~$\tau_1$ is further distant than the final \shortHorPlanner{} prediction time~$N t_\mathrm{d}$.
It holds that $\tau_1\geq N t_d\Leftrightarrow [\lambda_N=0]$, so the reachable set can be conditioned on~$\lambda_N$ by
\begin{equation}
	\label{eq:coupling_terminal}
[\lambda_N=0] \implies (\tau_1, \sigma_1)\in\reachspace{}(N t_d, s_N).
\end{equation}

The final \lstp{}, formulated as an~\ac{MIQP}, can be stated by decisions variables, costs, and constraints of the~\shortHorPlanner{}, the~\longHorPlanner{}, and with the additional coupling constraints~\eqref{eq:sh_lh_coupling_constraints} and~\eqref{eq:coupling_terminal}.

The \shortHorPlanner{} decision variables are~$X_\mathrm{s}=(X,U,\Lambda)\in V_\mathrm{s}$, where
	\begin{equation*}
		V_\mathrm{s}=
		 \R^{N\times n_x}\times
		 \R^{N-1\times n_u}\times
		 (\intSet[0]{1})^{N},
	\end{equation*}
and the \longHorPlanner{} decision variables are~$X_\mathrm{l}:=(\Sigma ,T,R,\Slack_{\mathrm{bi}}, B)\in V_\mathrm{l}$, where
	\begin{align*}
		\begin{split}
		V_\mathrm{l}:=
		\R^{L-1}\times
		\R^{L-1}\times
		\R^{L-1}\times
	\R^{L-1+N}\times
	(\intSet[0]{1})^{L-1\times M+2}.
		\end{split}
	\end{align*}

In total, $3(L-1)$ continuous and $(L-1) (M+2)$ binary variables are used to model the transitions for $LM$ \opps{}. Another $(N-1)(n_x+n_u) + n_x$ continuous and $N$ binary variables model the first lane change for a horizon of $t_\mathrm{d}N$. A total of $L+N-1$ variables are used as auxiliary variables.

Remarkably, the total number of binary variables is~$N_\mathrm{bin}=(L-1) (M+2)+N$, which is with~$O(LM+N)$ usually a much lower number in contrast to~$O(LMN)$ of~\cite{Quirynen2023}. 

\begin{figure}
	\begin{center}
		\includegraphics[trim={30mm 0mm 40mm 4mm},clip,width=90mm]{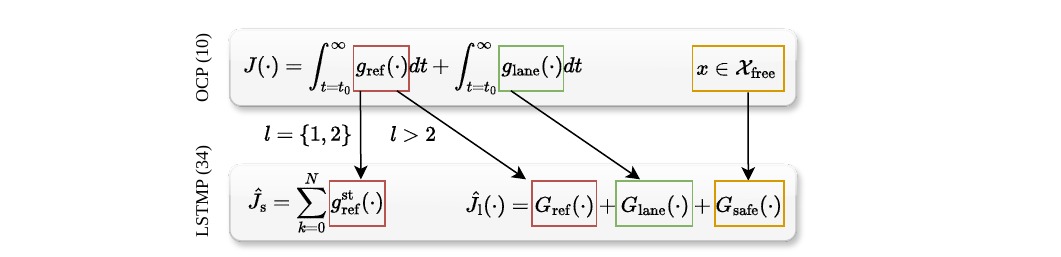}
		\caption{Overview of the approximations of the general \ac{OCP} objective function~\eqref{eq:ocp} by the \longHorPlanner{} cost~$\hat{J}_\mathrm{l}(\cdot)$ and the \shortHorPlanner{} cost~$\hat{J}_\mathrm{s}(\cdot)$. The reference cost~$g_\mathrm{ref}(\cdot)$ is approximated for the first two lanes within the~\shortHorPlanner{} and, thereafter, by the \longHorPlanner{}.}
		\label{fig:cost_relations}
	\end{center}
\end{figure}

The cost function~\eqref{eq:ocp_cost} is approximated by the cost of the \shortHorPlanner{} trajectory
\begin{equation*}
	\hat{J}_\mathrm{s}(X_\mathrm{s})=\sum_{k=0}^{N}
	g_\mathrm{ref}^\mathrm{st}(x_k,u_k,\lambda_k),
\end{equation*}
and the cost of the long horizon is
\begin{equation*}
	\hat{J}_\mathrm{l}(X_\mathrm{l})=G_\mathrm{lane}(\Sigma,T,B) +
	G_\mathrm{ref}(T,\Sigma) +
	G_\mathrm{safe}(R).
\end{equation*}
The relations of the general \ac{OCP} objective in~\eqref{eq:ocp} approximated by the \lstp{}, comprising the \longHorPlanner{} cost~$\hat{J}_\mathrm{l}(\cdot)$ and the \shortHorPlanner{} cost~$\hat{J}_\mathrm{s}(\cdot)$ are shown in Fig.~\ref{fig:cost_relations}.

Including a constraint $x_0= \hat{x}$ that constrains the decision variable~$x_0$ to the current state~$\hat{x}$, the constraints of the \shortHorPlanner{} are summarized by~$g_\mathrm{s}(X_\mathrm{s})\leq 0$ and include the discrete dynamic model~\eqref{eq:dynamics}, the control and state constraints~\eqref{eq:control_set} and the constraints related to the first lane-change~\eqref{eq:lane_change}, \eqref{eq:cost_sh_ref}, \eqref{eq:lane_change_bin} and \eqref{eq:lane_change_terminal}.

For the \longHorPlanner{}, a constraint~$g_\mathrm{l}(X_\mathrm{l})\leq 0$ summarizes the reachability constraints~\eqref{eq:reach}, the \ac{CC} constraints~\eqref{eq:cc_standard}, and the constraints used to formulate the disjunction among gaps in~\eqref{eq:disjunction_pre}, \eqref{eq:disjunction_post}, \eqref{eq:disjunction_bin1}, and \eqref{eq:disjunction_bin2}.

The coupling constraints~\eqref{eq:sh_lh_coupling_constraints} and \eqref{eq:coupling_terminal} between states of the \longHorPlanner{} and \shortHorPlanner{} are concisely written as~$g_\mathrm{c}(X_\mathrm{s},X_\mathrm{l})\leq 0$.

Ultimately, the \lstp{} approximates the solution of the \ac{OCP}~\eqref{eq:ocp} by solving the following \ac{MIQP} in each iteration
\begin{mini!}
	{\substack{X_\mathrm{s}\in V_\mathrm{s},\\X_\mathrm{l}\in V_\mathrm{l}}}			
	{\hat{J}_\mathrm{s}(X_\mathrm{s})+\hat{J}_\mathrm{l}(X_\mathrm{l})}
	{\label{eq:ocp_lstp}} 
	{} % result of optimization, e.g., J^* =
	%\addConstraint{LHS.1}{RHS.1\label{Const1}}{extraConst1}
	% \breakObjective{ + \norm{x^\mathrm{F}_N-x^\mathrm{F}_{\mathrm{ref},N}}_{Q_N}^2 }
	\addConstraint{g_\mathrm{s}(X_\mathrm{s})\leq 0,\quad}{g_\mathrm{l}(X_\mathrm{l})\leq 0,\quad}{g_\mathrm{c}(X_\mathrm{s},X_\mathrm{l})\leq 0}.
\end{mini!}
The output~$X^*=(X_\mathrm{s},X_\mathrm{l})$ of the~\lstp{} is always safe w.r.t. the obstacle constraints~\eqref{eq:free_3d_front} and~\eqref{eq:free_3d_lc}. This follows directly from the constraint formulations of the~\shortHorPlanner{}, including the terminal safe set~\eqref{eq:lane_change_terminal}. Approximation errors in the \longHorPlanner{} formulations may lead to sub-optimal behavior. However, they do not influence safety related to the feasibility of the trajectory~$X_\mathrm{s}^*$.

\section{Evaluation}
\label{sec:evaluation}
\begin{figure}
	\begin{center}
		\includegraphics[width=80mm]{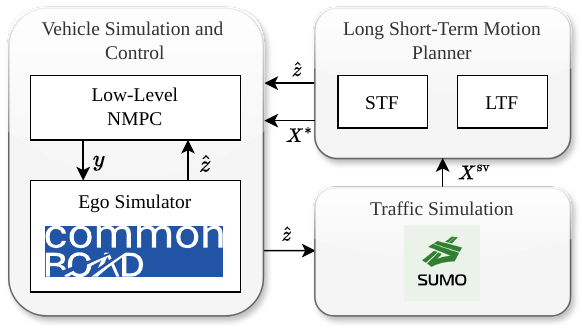}
		\caption{Overview of the adopted simulation architecture. After obtaining the ego vehicle state~$\hat{z}$ and the \opps{} states~$X^\mathrm{sv}$, the \lstp{} solves in each planning iteration the \ac{MIQP}~\eqref{eq:ocp_lstp}. The computed plan~$X^*$ related to the point-mass model is forwarded to the \nmpc{} for tracking. For the ego vehicle simulation, a \emph{BMW~320i} vehicle model provided by \texttt{CommonRoad} is used. The position of the ego vehicle~$\hat{z}$ is passed to the \texttt{SUMO} traffic simulator.}
		\label{fig:architecture}
	\end{center}
\end{figure}

We evaluate the proposed \lstp{} approach in two different setups. First, deterministic \opps{} are simulated as they are modeled in the \lstp{} and exact tracking of the provided plan is assumed.
A second setup includes more realistic scenarios, where the traffic is simulated interactively by the traffic simulator \texttt{SUMO}~\cite{Sumo2018}, based on benchmark scenarios provided by the \texttt{CommonRoad}-framework~\cite{Althoff2017}, cf., Fig.~\ref{fig:architecture}. Moreover, the \lstp{} is integrated into an \ac{AD}-stack with a \nmpc{} tracking controller of~\cite{Reiter2023a} that tracks the \lstp{} trajectory~$X^*$ by controlling a simulated single-track \emph{BMW~320i} medium-sized passenger car model provided by \texttt{CommonRoad}.
The \opp{} states~$\hat{X}^\mathrm{SV}$ and the current estimated point-mass state~$\hat{x}$ are the inputs of the planner. The point-mass state~$\hat{x}$ is obtained from the six-dimensional simulated single-track vehicle state~$\hat{z}$.

For both setups, the \lstp{} is compared against the \ac{MIP-DM} of~\cite{Quirynen2023} and a \astar{} formulation according to~\cite{Ajanovic2018}. 
Rendered simulations can be found on the website \url{https://rudolfreiter.github.io/lstmp_vis/}

\subsection{Implementation details}
We describe the setup used for evaluation in the following. Parameters are chosen according to Tab.~\ref{tab:params}.
\begin{table}[]
 	\caption{Parameter for evaluations.}
	\centering
	\begin{tabular}[t]{l r}
		\toprule
		\multicolumn{2}{c}{General parameters}\\
		\midrule
		$t_\mathrm{d}$, $M$ & $300$ms, 7\\
		$w_n,w_v,w_g,w_\mathrm{safe}$ & $10^{-2},10^{-1},200,10^{-5}$\\
		$R$ & diag$\big([5\cdot10^{-4},\; 2\cdot 10^{-3}]\big)$\\ 
		$\laneWidth$ & $3.75$m (Germany), $12$ feet (US)\\
		$[\lb{a}_\mathrm{lon}, \lb{a}_\mathrm{lat}]$ & $[-8,\;-3]\frac{\mathrm{m}}{\mathrm{s}^2}$\\
		$[\ub{a}_\mathrm{lon}, \ub{a}_\mathrm{lat}]$ & $[5,\;3]\frac{\mathrm{m}}{\mathrm{s}^2}$\\
		\midrule
		\multicolumn{2}{c}{\lstp{} - deterministic scenario}\\
		\midrule
		$t_\mathrm{d}$, $N$, $M$ & $300$ms, 15, 7\\
		$t_\mathrm{f}, \; \ub{t}_\mathrm{lc}$ & $10^5$s , $2.7$s\\
		$\ub{v}^\mathrm{sv},\lb{v}^\mathrm{sv}$ & $\hat{v}^\mathrm{sv},\hat{v}^\mathrm{sv}$\\
		\midrule
		\multicolumn{2}{c}{\lstp{}-V0 - interactive scenario}\\
		\midrule
		$\ub{v}^\mathrm{sv}$, $\lb{v}^\mathrm{sv}$, $\dimLanes{}$ & $\hat{v}^\mathrm{sv} +1\frac{\mathrm{m}}{\mathrm{s}}$, $\hat{v}^\mathrm{sv} -1\frac{\mathrm{m}}{\mathrm{s}}$, 5\\
		\midrule
		\multicolumn{2}{c}{\lstp{}-V1 - interactive scenario}\\
		\midrule
		$\ub{v}^\mathrm{sv}$, $\lb{v}^\mathrm{sv}$, $\dimLanes{}$ & $\hat{v}^\mathrm{sv} +1\frac{\mathrm{m}}{\mathrm{s}}$, $\hat{v}^\mathrm{sv} -1\frac{\mathrm{m}}{\mathrm{s}}$, 3\\
		\midrule
		\multicolumn{2}{c}{\lstp{}-V2 - interactive scenario}\\
		\midrule
		$\ub{v}^\mathrm{sv}$, $\lb{v}^\mathrm{sv}$, $\dimLanes{}$ & $\hat{v}^\mathrm{sv} + 3\frac{\mathrm{m}}{\mathrm{s}}$,
		$\hat{v}^\mathrm{sv} - 3\frac{\mathrm{m}}{\mathrm{s}}$, 5\\
		\bottomrule 
		\addlinespace
	\end{tabular}
	\label{tab:params}
\end{table}

\subsubsection{Preprocessing}
The \opps{} states~$X^\mathrm{SV}$ are processed before either planner is executed. First, \opps{} that drive closer to each other than a longitudinal threshold distance of~$15\mathrm{m}$ are merged by setting the corresponding upper and lower velocity bounds and increasing the occupied space. Second, a maximum number of~$M=7$~\opps{} per lane are considered, which are the~$7$ closest vehicles at the current time step on each lane.

\subsubsection{Benchmark \ac{MIP-DM}}
The first benchmark is based on the \ac{MIP-DM} formulation of~\cite{Quirynen2023}. It uses a fixed discrete-time trajectory, similar to the \shortHorPlanner{}, however, with four binary variables per obstacle and per time step to account for the rectangular obstacle shape. One further binary variable per time step indicates a lane change. The number of binary variables for the \ac{MIP-DM} is therefore~$N_\mathrm{bin}=4NML+N$. The \ac{MIP-DM} is adapted to be comparable to the \lstp{}. First, only one lane change direction is allowed, which reduces the number of binary variables. Second, obstacle shapes are inflated to occupy the whole lane, equally to the \lstp{}. Finally, the interactive braking behavior of succeeding \opps{} on the same lane is implemented by deactivating corresponding obstacles on the current lane at the current time step. For the \ac{MIP-DM} a total number of $M=3$ vehicles are considered on $L=5$ consecutive lanes, while the horizon length~$N$ is $10,15$~or~$20$ steps.

\subsubsection{Benchmark \astar{}}
The second benchmark is based on the \astar{} of~\cite{Ajanovic2018}. This planner considers lateral motion only at the discrete lane indices, with a search space of $(t,s,l)$. In order to be comparable to the other planners, we modify the search space to $(s,l,v_s)$, which includes the velocity~$v_s$ instead of time. Since the \astar{} of \cite{Ajanovic2018} does not consider lateral states between lane centers, we use a sampling time of $7 t_\mathrm{d}$ to allow full lane changes in one expansion, i.e., it is guaranteed that the final planning vertex is always located on the center of a lane. We use the same planner model~\eqref{eq:dynamics} for vertex expansions. Note that \astar{} could use nonlinear models without increasing the computation time, which, in contrast, would be challenging for the \lstp{} and \ac{MIP-DM}. As an admissible heuristic, the relaxed solution of~\eqref{eq:ocp_lstp} without obstacle constraints is computed for each lane. The longitudinal acceleration control is discretized into~$11$ intervals and the lateral acceleration is computed by using~$11$~lane change primitives. The lateral states correspond to the number of lanes, the longitudinal position is discretized with~$100$ intervals, and the velocity with~$20$ intervals. The number of node expansions is varied in experiments between~$5$ and~$500$.

\subsubsection{Low-level NMPC}
The \nmpc{} is formulated as shown in~\cite{Reiter2023a}, using a nonlinear single-track vehicle model, a sampling time of~$10\mathrm{ms}$ and a horizon of~$1.5\mathrm{s}$.
The controls comprise the acceleration~$a$ and the steering rate~$\dot\delta$.

\subsubsection{Scenarios}
\begin{figure}
	\begin{center}
		\includegraphics[scale=0.7]{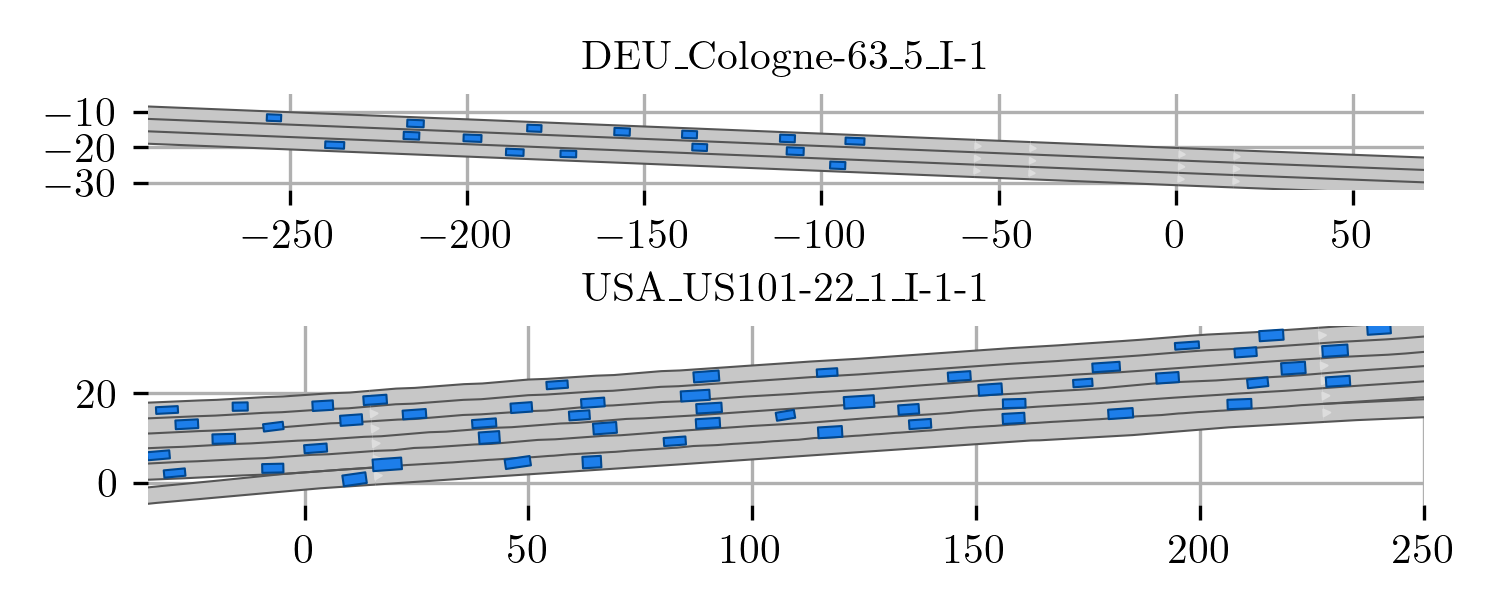}
		\caption{Different tracks from the \texttt{CommonRoad} scenario database used for the closed-loop simulation.}
		\label{fig:tracks}
	\end{center}
\end{figure}

For deterministic comparisons in Sect.~\ref{sec:eval_deterministic} and interactive closed-loop comparisons in Sect.~\ref{sec:eval_interactive}, the scenarios are chosen according to Tab.~\ref{tab:scenario_settings}. Due to traffic congestion, the velocity can be zero. Traffic flow and density are averaged over the simulation. The velocity range~$V^\mathrm{sv}$ corresponds to the measured \opps{} velocities during all simulations. 
\begin{table}
\caption{Different scenario settings for \opps{} used in evaluations.}
	\centering
	\ra{1.2}
	\begin{tabular}{@{}l|ccccc@{}}
		\addlinespace
		\toprule
		Scenario Name	  & tr.-flow  & tr.-density & $L$ & $V^\mathrm{sv}$ &  $\refer{v}$\\
		\midrule
		  & $\frac{\mathrm{\opps{}}}{\mathrm{lane}\cdot \mathrm{min}}$  & $\frac{\mathrm{\opps{}}}{\mathrm{lane}\cdot \mathrm{km}}$ &  & $\frac{\mathrm{m}}{\mathrm{s}}$ & $\frac{\mathrm{m}}{\mathrm{s}}$\\
		\midrule
		\midrule
		 \multicolumn{6}{c}{ Deterministic}\\
		\midrule
		custom & 14.6 & 12.2 & 9 &$[15,  35]$ & 25\\
		\midrule
		\multicolumn{6}{c}{Closed-loop interactive }\\
		\midrule
		USA\_US101-22\_1\_I-1 & 11.8 & 14.0 & 6 & $[0 , 22.2]$ & 15\\
		DEU\_Col.-63\_5\_I-1 & 22.3 & 26.9 & 3 & $[11.0 , 16.5]$& 11\\
		DEU\_Col.-63\_5\_I-1\_s & 8.9 & 10.5 & 3 & $[11.1 , 16.5]$& 15\\
		\bottomrule
		\addlinespace
	\end{tabular}
	\label{tab:scenario_settings}
\end{table}
The scenarios are simulated for~$40$ seconds or until the end of the road is reached.
Snapshots of the \texttt{CommonRoad} scenarios for interactive simulations are shown in~Fig.~\ref{fig:tracks}.
\subsubsection{Computations and numerical solvers}
The \acp{MIQP} of the \lstp{} and the \ac{MIP-DM} are solved with \texttt{gurobi}~\cite{Gurobi2023}.
The \ac{NLP}, arising in the \nmpc{}, is solved by the open-source solver \texttt{acados}~\cite{Acados2021}.
Simulations are executed on a \texttt{LENOVO ThinkPad L15 Gen 1} Laptop with an \texttt{Intel(R) Core(TM) i7-10510U @ 1.80GHz} CPU.

\subsection{Evaluation for deterministic traffic and exact tracking}
\label{sec:eval_deterministic}
\begin{figure}
	\begin{center}
		\includegraphics[trim={0cm 0mm 0cm 3mm},clip,width=85mm]{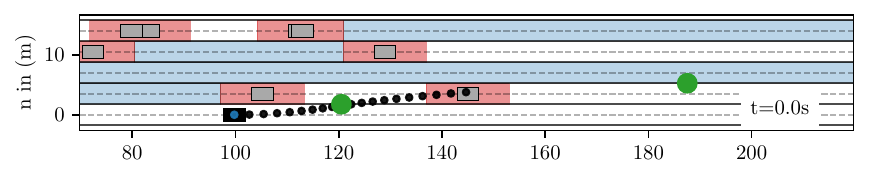}
		\includegraphics[trim={0cm 0mm 0cm 3mm},clip,width=85mm]{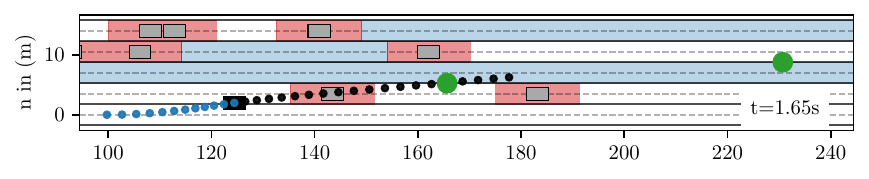}
		\includegraphics[trim={0cm 0mm 0cm 3mm},clip,width=85mm]{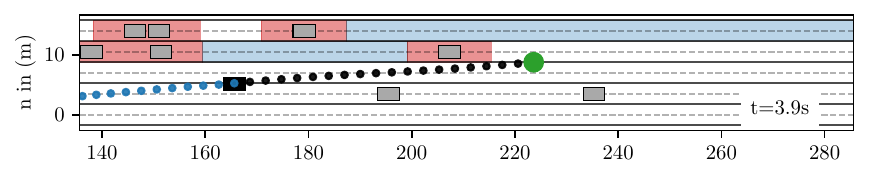}
		\includegraphics[trim={0cm 0mm 0cm 3mm},clip,width=85mm]{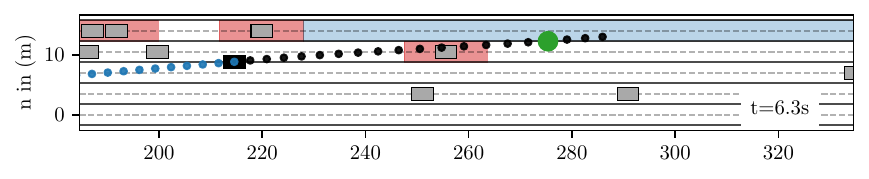}
		\includegraphics[trim={0cm 3mm 0cm 3mm},clip,width=85mm]{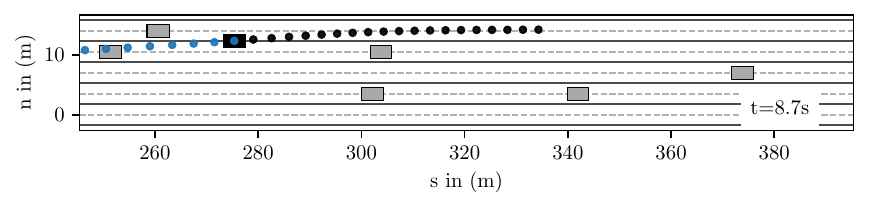}
		\caption{Snapshots during lane-changes on a five-lane deterministic environment with randomized \opp{} (gray) initial speeds and lanes. Blue regions indicate computed gaps of the \lstp{}, with green points corresponding to the expected transition position~$\sigma_l$ and the black \shortHorPlanner{} trajectory~$X^*$. Red areas correspond to occupied sets~$\occspace{}(t_\mathrm{sim})$, where~$t_\mathrm{sim}$ is the current simulation time of the snapshot.}
		\label{fig:lane_change_snapshots}
	\end{center}
\end{figure}

In order to compare the performance of the planner without interference from the traffic prediction error, simulation modeling error, and controller performance, the planner is simulated with deterministic \opps{} and exact tracking.
Deterministic \opps{} are simulated with constant speed, if they are at a minimum distance to a slower leading \opps{} and with the speed of the leading \opp{} if they are below the threshold distance. 
The planned trajectory~$X^*$ is assumed to be tracked exactly.
This setup resembles the model of the traffic used in the \lstp{}, where tight bounds for the obstacle-free sets~\eqref{eq:free_3d_front} and~$\eqref{eq:free_3d_lc}$ can easily be found.
In Fig.~\ref{fig:lane_change_snapshots}, snapshots of a randomized simulation with five lanes are shown, where the vehicle starts at the bottom lane and has to reach the top lane.
Red areas indicate the \opps{} after pre-processing. The \shortHorPlanner{} trajectory~$X^*$ of the \lstp{} is shown in black, whereas the transition gaps are shown in blue.
\begin{figure}
	\begin{center}
		\includegraphics[scale=0.85]{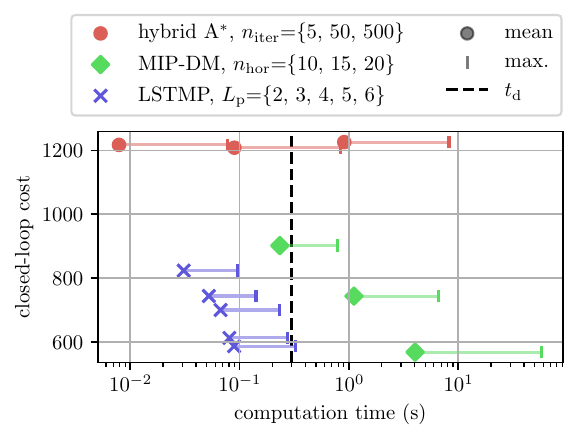}
		\caption{Pareto comparison of planners in randomized deterministic traffic scenarios. The \astar{} planner can be parameterized to have the fastest computation time and the \ac{MIP-DM} achieves the lowest costs. However, the novel \lstp{} formulation performs best when both, a low computation time as well as low costs are required.}
		\label{fig:comp_planners_offline}
	\end{center}
\end{figure}
\begin{figure*}
	\begin{center}
		\includegraphics[scale=0.7]{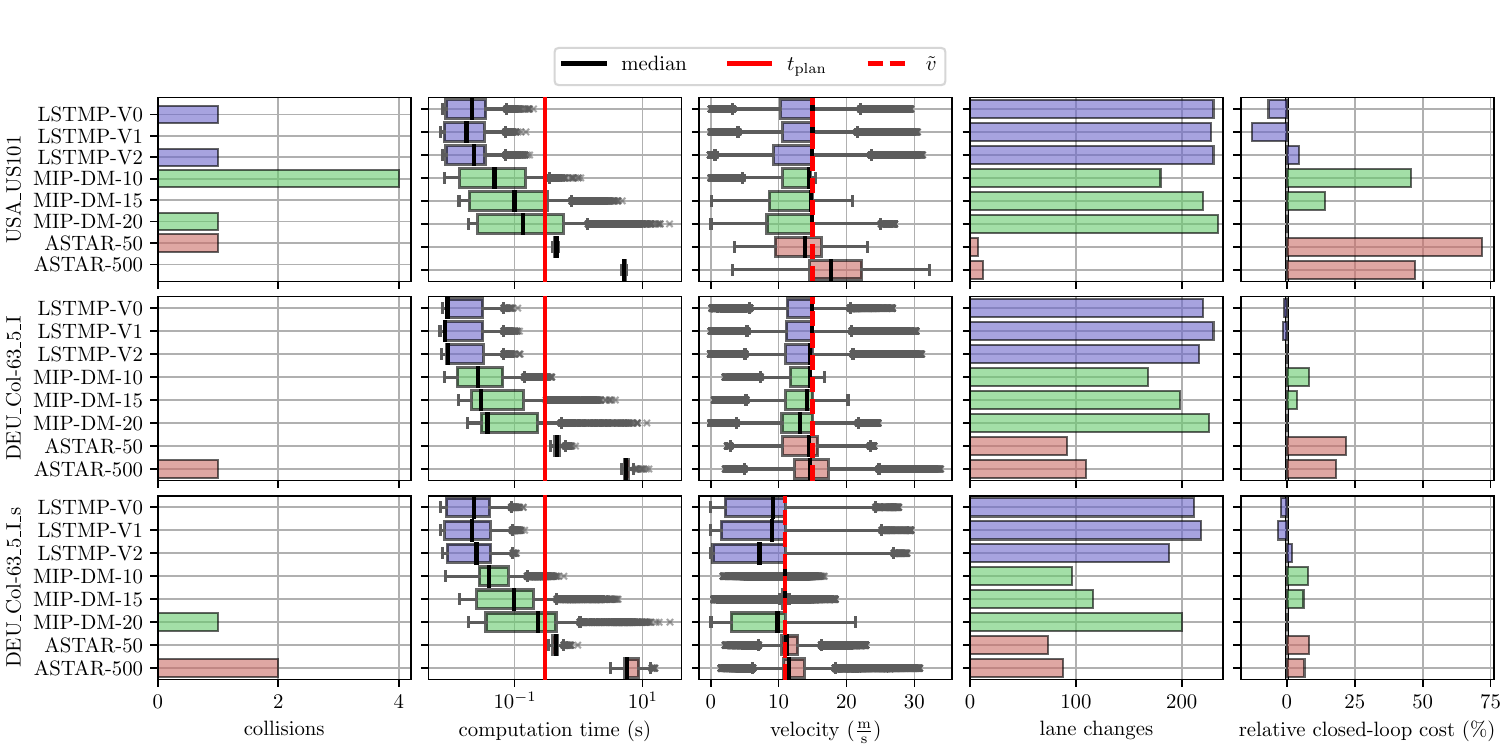}
		\caption{Closed-loop evaluation of variants of the proposed \lstp{} planner (blue) compared to \astar{} (red) and \ac{MIP-DM} (green). The total number of collisions, the computation time, velocities, the total number of lane changes, and the closed-loop cost compared to the most exact \ac{MIP-DM} formulation are compared for different randomized scenarios. The proposed \lstp{} has the lowest computation time below the planning time~$t_\mathrm{plan}$ and a high number of lane changes for all scenarios. Moreover, it can reduce the closed-loop cost significantly. Occasional collisions are observed for all variants, irrespective of their parameters.}
		\label{fig:costs}
	\end{center}
\end{figure*}
The evaluated closed-loop cost and computation time of~$100$ randomized \emph{custom} scenarios according to~Tab.~\ref{tab:scenario_settings} for the deterministic setup are shown in Fig.~\ref{fig:comp_planners_offline} on the Pareto front.
	
The comparisons include evaluations for different parameter settings of the algorithms, i.e., the number of considered consecutive lanes in the \lstp{}, the maximum node expansions in~\astar{}, and the horizon length of the~\ac{MIP-DM}.

The \ac{MIP-DM} with the longest horizon outperforms the \lstp{} in the average closed-loop cost over the full simulations, however, at a high computational expense which violates the real-time requirement.
In fact, the computation time of the \ac{MIP-DM} is an order of magnitude higher than of the \lstp{}.

The \astar{} can be faster to execute compared to the the \lstp{}, but it yields a higher closed-loop cost. In our experiments, increasing the iterations of \astar{} could not yield a better performance. This may be due to the longer duration of motion primitives to allow lane changes and the resulting coarser time discretization.
\begin{table}
	\caption{Comparison of planners in randomized deterministic traffic scenarios for different mean (maximum) quantities.}
	\centering
	\ra{1.2}
	\begin{tabular}{@{}lcc|cccc@{}}
		\addlinespace
		\toprule
		\multicolumn{3}{c}{Planner}	  & $\Delta \tilde{v}$  & $a_\mathrm{lat}$ & $a_\mathrm{lon}$ &  $l_\mathrm{max}$  \\
		\midrule
		Par.	 & Val.	  &$N_\mathrm{bin}$ & $\frac{\mathrm{m}}{\mathrm{s}}$  & $10^{\text{-}1}\frac{\mathrm{m}}{\mathrm{s}^2}$ & $10^{\text{-}1}\frac{\mathrm{m}}{\mathrm{s}^2}$ & \\
		\midrule
		\multicolumn{3}{c}{\lstp{}} & \multicolumn{4}{c}{ }\\
		\midrule	
		$L_\mathrm{p}$	& 2 &22& 1.75 & 0.71 (4.58) & 1.40 (5.30) & 3.92 \\
						& 3 &29& 0.42 & 0.63 (3.82) & 1.46 (6.35) & 4.20 \\
						& 4 &36& 0.03 & 0.66 (3.70) & 1.48 (6.45) & 4.34 \\
						& 5 &43& 0.07 & 0.64 (3.62) & 1.37 (4.96) & 4.62 \\
						& 6 &50& 0.03 & 0.67 (3.62) & 1.55 (6.15) & 4.66 \\
		\midrule
		\multicolumn{3}{c}{\ac{MIP-DM}}& \multicolumn{4}{c}{ } \\
		\midrule 
		$N$	 			& 10 &610& 2.50 & 0.48 (2.68) & 0.69 (5.14) & 3.62 \\
						& 15 &915& 2.10 & 0.52 (2.67) & 1.06 (8.16) & 4.17 \\
						& 20 &1220& 1.98 & 0.65 (2.67) & 1.83 (8.70) & 4.75 \\
		\midrule
		\multicolumn{3}{c}{\astar{}} & \multicolumn{4}{c}{ }\\
		\midrule
		iter.	 		& 5 &N/A& 1.78 & 0.47 (2.40) & 0.14 (1.00) & 2.61 \\
						& 50 &N/A& 1.73 & 0.47 (2.40) & 0.17 (1.14) & 2.63 \\
						& 500 &N/A& 1.70 & 0.43 (2.40) & 0.22 (1.43) & 2.58 \\
		\bottomrule
		\vspace{0.5px}
	\end{tabular}
	\label{tab:comparion_planners_offline}
\end{table}
Further relevant properties related to the lane change multi-objective~\eqref{eq:ocp_cost} are shown in Tab.~\ref{tab:comparion_planners_offline}. This includes the mean deviation from the reference speed~$\Delta \tilde{v}$, mean and maximum values for the lateral and longitudinal accelerations, and the maximum reached lane~$l_\mathrm{max}$ at the end of the simulation. It shows that the \lstp{} with a longer horizon better keeps the reference speed and also changes lanes more often. By utilizing large acceleration values, the \ac{MIP-DM} achieves the highest number of lane changes and the overall lowest closed-loop cost, c.f., Fig.~\ref{fig:comp_planners_offline}.

Notably, the number of binary variables required in the \ac{MIP-DM} is much larger than in the \lstp{}, which leads to a significantly longer computation time. For a prediction horizon of~$N=10$ and settings of Tab.~\ref{tab:params}, the \ac{MIP-DM} requires~$610$ binary variables and for a prediction horizon of~$N=20$ a total of~$1220$ binary variables. The \lstp{} that considers in total $L_\mathrm{p}=2$ lanes, requires only~$22$ binary variables, whereas considering $L_\mathrm{p}=5$ lanes requires only~$50$ binary variables.

\subsection{Evaluation for interactive traffic and closed-loop control}
\label{sec:eval_interactive}
For different randomized scenarios according to Tab.~\ref{tab:scenario_settings} and Fig~\ref{fig:tracks}, the lane changing problem is simulated with interactive \opps{}, using a software architecture corresponding to Fig.~\ref{fig:architecture}.
The ego vehicle starts at random free positions and has to reach the leftmost lane, according to cost~\eqref{eq:ocp_cost}, with parameters of Tab.~\ref{tab:params}.
The \lstp{}, \astar{}, and \ac{MIP-DM} are compared with a low-level tracking controller in closed-loop simulations. States are assumed to be estimated exactly, however, the velocity range of \opps{} is unknown. 
Different settings of the planners are used according to Tab.~\ref{tab:params} to create the statistical evaluation of performance measures as shown in Fig.~\ref{fig:costs}.

The performance evaluations show rare collisions of all planners due to prediction errors. For the conservative \lstp{}-V1 configuration, no collisions were recorded.
Computation times are lowest for the \lstp{} planner and well below the planning time threshold~$t_\mathrm{plan}$. The computation times for the \astar{} are nearly constant since the planning nodes are expanded with a fixed number of iterations. Notably, the computations for \astar{} were not performed on a runtime-optimized code.
The velocity varies the most for the \lstp{}, which promotes acceleration and deceleration to reach certain gaps. This can also be verified by the high number of lane transitions of the \lstp{}. Particularly, in the DEU\_Col.-63\_5\_I-1\_s scenario long-term decisions significantly raised the number of lane transitions in the~\lstp{}.
The closed-loop cost for~\lstp{} configurations are below the benchmark comparisons, particularly, below the \ac{MIP-DM}-20 with the longest horizon of~$20$~steps, which we define as \emph{expert}. Costs are relatively expressed to the cost of \ac{MIP-DM}-20 and outperformed by \lstp{}-V0 and \lstp{}-V1.
\section{Conclusion and Discussion}
\label{sec:conclusion}
\acresetall
Under the variety of different planning methods for \ac{AD}, the proposed \lstp{} for lane change planning achieves a good trade-off between performance and computational costs, thanks to the use of state-of-the-art \ac{MIQP} solvers. 
The considered problem has relevant combinatorial and continuous parts, which makes \ac{MIQP} formulations particularly suited to solve the proposed motion planning problem. 
Building on previous work to minimize the number of combinatorial variables, we introduced a novel long-horizon approximation. 
Together with a discrete-time trajectory, a single \ac{MIQP}, which is computationally very efficient, was formulated consistently.

We compared our approach to the~\ac{MIP-DM}~\cite{Quirynen2023}, which uses more integer variables to model rectangular obstacle shapes that are not required to be aligned with the lane boundaries and to model lane transitions in both directions. This makes the \ac{MIP-DM} a more versatile approach, i.e., lane changing is only a subset of problems that can be addressed with it.

The fundamental modeling approach of the \ac{LSTMP} is the decomposition into convex cells together with a simplification due to the road alignment. The authors assume that is possible to add integer variables to achieve lane transitioning in both directions for a fixed maximum number of transitions and additional convex decompositions to resemble nonconvexities in the \st{}, as for instance traffic lights. 
	
In the future we will evaluate whether more flexible mixed-integer nonlinear programming can achieve better performance under real-time requirements.

\bibliographystyle{IEEEtran}
%\newpage
\bibliography{lib} 

\begin{thebibliography}{10}
\providecommand{\url}[1]{#1}
\csname url@samestyle\endcsname
\providecommand{\newblock}{\relax}
\providecommand{\bibinfo}[2]{#2}
\providecommand{\BIBentrySTDinterwordspacing}{\spaceskip=0pt\relax}
\providecommand{\BIBentryALTinterwordstretchfactor}{4}
\providecommand{\BIBentryALTinterwordspacing}{\spaceskip=\fontdimen2\font plus
\BIBentryALTinterwordstretchfactor\fontdimen3\font minus
  \fontdimen4\font\relax}
\providecommand{\BIBforeignlanguage}[2]{{%
\expandafter\ifx\csname l@#1\endcsname\relax
\typeout{** WARNING: IEEEtran.bst: No hyphenation pattern has been}%
\typeout{** loaded for the language `#1'. Using the pattern for}%
\typeout{** the default language instead.}%
\else
\language=\csname l@#1\endcsname
\fi
#2}}
\providecommand{\BIBdecl}{\relax}
\BIBdecl

\bibitem{Reif1979}
J.~H. Reif, ``Complexity of the mover's problem and generalizations,'' in
  \emph{20th Annual Symposium on Foundations of Computer Science (sfcs 1979)},
  1979, pp. 421--427.

\bibitem{LaValle2006}
S.~M. LaValle, \emph{Planning Algorithms}.\hskip 1em plus 0.5em minus
  0.4em\relax USA: Cambridge University Press, 2006.

\bibitem{Quirynen2023}
R.~Quirynen, S.~Safaoui, and S.~{\relax Di}~Cairano, ``Real-time mixed-integer
  quadratic programming for vehicle decision making and motion planning,''
  \emph{ArXiv}, vol. abs/2308.10069, 2023.

\bibitem{Miller2018}
C.~Miller, C.~Pek, and M.~Althoff, ``Efficient mixed-integer programming for
  longitudinal and lateral motion planning of autonomous vehicles,'' in
  \emph{IEEE Intelligent Vehicles Symposium (IV)}, 2018, pp. 1954--1961.

\bibitem{Althoff2017}
M.~Althoff, M.~Koschi, and S.~Manzinger, ``Commonroad: Composable benchmarks
  for motion planning on roads,'' in \emph{IEEE Intelligent Vehicles Symposium
  (IV)}, 2017, pp. 719--726.

\bibitem{Paden2016}
B.~{Paden}, M.~{Čáp}, S.~Z. {Yong}, D.~{Yershov}, and E.~{Frazzoli}, ``A
  survey of motion planning and control techniques for self-driving urban
  vehicles,'' \emph{IEEE Transactions on Intelligent Vehicles}, vol.~1, no.~1,
  pp. 33--55, 2016.

\bibitem{Claussmann2020}
L.~Claussmann, M.~Revilloud, D.~Gruyer, and S.~Glaser, ``A review of motion
  planning for highway autonomous driving,'' \emph{IEEE Transactions on
  Intelligent Transportation Systems}, vol.~21, no.~5, pp. 1826--1848, 2020.

\bibitem{Reda2024}
M.~Reda, A.~Onsy, A.~Y. Haikal, and A.~Ghanbari, ``Path planning algorithms in
  the autonomous driving system: A comprehensive review,'' \emph{Robotics and
  Autonomous Systems}, vol. 174, p. 104630, 2024.

\bibitem{Szilard2022}
S.~Aradi, ``Survey of deep reinforcement learning for motion planning of
  autonomous vehicles,'' \emph{IEEE Transactions on Intelligent Transportation
  Systems}, vol.~23, no.~2, pp. 740--759, 2022.

\bibitem{Ravi2022}
B.~R. Kiran, I.~Sobh, V.~Talpaert, P.~Mannion, A.~A.~A. Sallab, S.~Yogamani,
  and P.~Pérez, ``Deep reinforcement learning for autonomous driving: A
  survey,'' \emph{IEEE Transactions on Intelligent Transportation Systems},
  vol.~23, no.~6, pp. 4909--4926, 2022.

\bibitem{Mouhagir2016}
H.~Mouhagir, R.~Talj, V.~Cherfaoui, F.~Aioun, and F.~Guillemard, ``Integrating
  safety distances with trajectory planning by modifying the occupancy grid for
  autonomous vehicle navigation,'' in \emph{IEEE 19th International Conference
  on Intelligent Transportation Systems (ITSC)}, 2016, pp. 1114--1119.

\bibitem{Sheckells2017}
M.~Sheckells, T.~M. Caldwell, and M.~Kobilarov, ``Fast approximate path
  coordinate motion primitives for autonomous driving,'' in \emph{IEEE 56th
  Annual Conference on Decision and Control (CDC)}, 2017, pp. 837--842.

\bibitem{Klancar2021}
G.~Klančar, M.~Seder, S.~Blažič, I.~Škrjanc, and I.~Petrović, ``Drivable
  path planning using hybrid search algorithm based on e* and
  bernstein–bézier motion primitives,'' \emph{IEEE Transactions on Systems,
  Man, and Cybernetics: Systems}, vol.~51, no.~8, pp. 4868--4882, 2021.

\bibitem{Kavraki1996}
L.~Kavraki, P.~Svestka, J.-C. Latombe, and M.~Overmars, ``Probabilistic
  roadmaps for path planning in high-dimensional configuration spaces,''
  \emph{IEEE Transactions on Robotics and Automation}, vol.~12, no.~4, pp.
  566--580, 1996.

\bibitem{Karaman2011}
S.~Karaman and E.~Frazzoli, ``Sampling-based algorithms for optimal motion
  planning,'' \emph{Int. J. Rob. Res.}, vol.~30, no.~7, p. 846–894, jun 2011.

\bibitem{Oktay2017}
O.~Arslan, K.~Berntorp, and P.~Tsiotras, ``Sampling-based algorithms for
  optimal motion planning using closed-loop prediction,'' in \emph{IEEE
  International Conference on Robotics and Automation (ICRA)}, 2017, pp.
  4991--4996.

\bibitem{Wang2020a}
J.~Wang, W.~Chi, C.~Li, C.~Wang, and M.~Q.-H. Meng, ``Neural {RRT}*:
  Learning-based optimal path planning,'' \emph{IEEE Transactions on Automation
  Science and Engineering}, vol.~17, no.~4, pp. 1748--1758, 2020.

\bibitem{Wang2021}
J.~Wang, B.~Li, and M.~Q.-H. Meng, ``Kinematic constrained bi-directional {RRT}
  with efficient branch pruning for robot path planning,'' \emph{Expert Systems
  with Applications}, vol. 170, p. 114541, 2021.

\bibitem{Montemerlo2008}
M.~Montemerlo, J.~Becker, S.~Bhat, H.~Dahlkamp, D.~Dolgov, S.~Ettinger,
  D.~Haehnel, T.~Hilden, G.~Hoffmann, B.~Huhnke, D.~Johnston, S.~Klumpp,
  D.~Langer, A.~Levandowski, J.~Levinson, J.~Marcil, D.~Orenstein, J.~Paefgen,
  I.~Penny, A.~Petrovskaya, M.~Pflueger, G.~Stanek, D.~Stavens, A.~Vogt, and
  S.~Thrun, ``Junior: The {S}tanford entry in the urban challenge,''
  \emph{Journal of Field Robotics}, vol.~25, no.~9, pp. 569--597, 2008.

\bibitem{Ajanovic2018}
Z.~Ajanovic, B.~Lacevic, B.~Shyrokau, M.~Stolz, and M.~Horn, ``Search-based
  optimal motion planning for automated driving,'' in \emph{IEEE/RSJ
  International Conference on Intelligent Robots and Systems (IROS)}, 2018, pp.
  4523--4530.

\bibitem{Dongchan2022}
D.~Kim, G.~Kim, H.~Kim, and K.~Huh, ``A hierarchical motion planning framework
  for autonomous driving in structured highway environments,'' \emph{IEEE
  Access}, vol.~10, pp. 20\,102--20\,117, 2022.

\bibitem{Jikun2020}
J.~Rong, S.~Arrigoni, N.~Luan, and F.~Braghin, ``Attention-based sampling
  distribution for motion planning in autonomous driving,'' in \emph{39th
  Chinese Control Conference (CCC)}, 2020, pp. 5671--5676.

\bibitem{Diehl2005}
M.~Diehl, H.~G. Bock, and J.~P. Schl\"{o}der, ``A real-time iteration scheme
  for nonlinear optimization in optimal feedback control,'' \emph{SIAM Journal
  on Control and Optimization}, vol.~43, no.~5, pp. 1714--1736, 2005.

\bibitem{Xiangjun2016}
X.~Qian, F.~Altché, P.~Bender, C.~Stiller, and A.~de~La~Fortelle, ``Optimal
  trajectory planning for autonomous driving integrating logical constraints:
  An miqp perspective,'' in \emph{IEEE 19th International Conference on
  Intelligent Transportation Systems (ITSC)}, 2016, pp. 205--210.

\bibitem{Li2022}
J.~Li, X.~Xie, Q.~Lin, J.~He, and J.~M. Dolan, ``Motion planning by search in
  derivative space and convex optimization with enlarged solution space,'' in
  \emph{IEEE/RSJ International Conference on Intelligent Robots and Systems
  (IROS)}, 2022, pp. 13\,500--13\,507.

\bibitem{Gurobi2023}
\BIBentryALTinterwordspacing
{Gurobi Optimization, LLC}, ``{Gurobi Optimizer Reference Manual},'' 2023.
  [Online]. Available: \url{https://www.gurobi.com}
\BIBentrySTDinterwordspacing

\bibitem{Johnson2013}
J.~Johnson and K.~Hauser, ``Optimal longitudinal control planning with moving
  obstacles,'' in \emph{IEEE Intelligent Vehicles Symposium (IV)}, 2013, pp.
  605--611.

\bibitem{Bender2015}
P.~Bender, O.~S. Tas, J.~Ziegler, and C.~Stiller, ``The combinatorial aspect of
  motion planning: Maneuver variants in structured environments,'' in
  \emph{IEEE Intelligent Vehicles Symposium (IV)}, 2015, pp. 1386--1392.

\bibitem{Marcucci2022}
T.~Marcucci, M.~Petersen, D.~von Wrangel, and R.~Tedrake, ``Motion planning
  around obstacles with convex optimization,'' \emph{Science Robotics}, vol.~8,
  no.~84, 2023.

\bibitem{Deolasee23}
S.~Deolasee, Q.~Lin, J.~Li, and J.~M. Dolan, ``Spatio-temporal motion planning
  for autonomous vehicles with trapezoidal prism corridors and {B}{\'{e}}zier
  curves,'' in \emph{American Control Conference, San Diego, CA, USA}.\hskip
  1em plus 0.5em minus 0.4em\relax {IEEE}, 2023, pp. 3207--3214.

\bibitem{Laurense2022}
V.~A. Laurense and J.~C. Gerdes, ``Long-horizon vehicle motion planning and
  control through serially cascaded model complexity,'' \emph{IEEE Transactions
  on Control Systems Technology}, vol.~30, no.~1, pp. 166--179, 2022.

\bibitem{Wang2021a}
P.~Wang, D.~Liu, J.~Chen, H.~Li, and C.-Y. Chan, ``Decision making for
  autonomous driving via augmented adversarial inverse reinforcement
  learning,'' in \emph{2021 IEEE International Conference on Robotics and
  Automation (ICRA)}, 2021, pp. 1036--1042.

\bibitem{Bronstein2022}
E.~Bronstein, M.~Palatucci, D.~Notz, B.~White, A.~Kuefler, Y.~Lu, S.~Paul,
  P.~Nikdel, P.~Mougin, H.~Chen, J.~Fu, A.~Abrams, P.~Shah, E.~Racah,
  B.~Frenkel, S.~Whiteson, and D.~Anguelov, ``Hierarchical model-based
  imitation learning for planning in autonomous driving,'' in \emph{IEEE/RSJ
  International Conference on Intelligent Robots and Systems (IROS)}, 2022, pp.
  8652--8659.

\bibitem{Szalay2018}
Z.~Szalay, T.~Tettamanti, D.~Esztergár-Kiss, I.~Varga, and C.~Bartolini,
  ``Development of a test track for driverless cars: Vehicle design, track
  configuration, and liability considerations,'' \emph{Periodica Polytechnica
  Transportation Engineering}, vol.~46, no.~1, p. 29–35, 2018.

\bibitem{Russell2020}
S.~Russell and P.~Norvig, \emph{Artificial Intelligence: {A} Modern Approach
  (4th Edition)}.\hskip 1em plus 0.5em minus 0.4em\relax Pearson, 2020.

\bibitem{Torrisi2004}
F.~Torrisi and A.~Bemporad, ``Hysdel-a tool for generating computational hybrid
  models for analysis and synthesis problems,'' \emph{IEEE Transactions on
  Control Systems Technology}, vol.~12, no.~2, pp. 235--249, 2004.

\bibitem{Williams2013}
H.~P. Williams, \emph{Model {B}uilding in {M}athematical {P}rogramming}.\hskip
  1em plus 0.5em minus 0.4em\relax Hoboken, N.J.: Wiley, 2013.

\bibitem{Boyd2004}
S.~Boyd and L.~Vandenberghe, \emph{Convex Optimization}.\hskip 1em plus 0.5em
  minus 0.4em\relax Cambridge University Press, 2004.

\bibitem{Zhou2023}
J.~Zhou, ``Interaction and uncertainty-aware motion planning for autonomous
  vehicles using model predictive control,'' Ph.D. dissertation, Linköping
  University Electronic Press, 2023.

\bibitem{Ziegler2014a}
J.~Ziegler, P.~Bender, T.~Dang, and C.~Stiller, ``Trajectory planning for
  {B}ertha - a local, continuous method,'' in \emph{IEEE Intelligent Vehicles
  Symposium, Proceedings}, 06 2014, pp. 450--457.

\bibitem{Werling2010a}
M.~Werling, J.~Ziegler, S.~Kammel, and S.~Thrun, ``Optimal trajectory
  generation for dynamic street scenarios in a {F}renet frame,'' in \emph{IEEE
  International Conference on Robotics and Automation}, 06 2010, pp. 987 --
  993.

\bibitem{Reiter2023a}
R.~Reiter, A.~Nurkanović, J.~Frey, and M.~Diehl, ``Frenet-{C}artesian model
  representations for automotive obstacle avoidance within nonlinear {MPC},''
  \emph{European Journal of Control}, p. 100847, 2023.

\bibitem{Eilbrecht2020}
J.~Eilbrecht and O.~Stursberg, ``Challenges of trajectory planning with
  integrator models on curved roads,'' \emph{IFAC-PapersOnLine}, vol.~53,
  no.~2, pp. 15\,588--15\,595, 2020, 21st IFAC World Congress.

\bibitem{Wang2020}
M.~Wang, N.~Mehr, A.~Gaidon, and M.~Schwager, ``Game-theoretic planning for
  risk-aware interactive agents,'' in \emph{IEEE/RSJ International Conference
  on Intelligent Robots and Systems (IROS)}, 2020, pp. 6998--7005.

\bibitem{Cleac2022}
S.~Le~Cleac, M.~Schwager, and Z.~Manchester, ``Algames: A fast augmented
  {L}agrangian solver for constrained dynamic games,'' \emph{Auton. Robots},
  vol.~46, no.~1, p. 201–215, jan 2022.

\bibitem{Burger2018}
C.~Burger and M.~Lauer, ``Cooperative multiple vehicle trajectory planning
  using miqp,'' in \emph{21st International Conference on Intelligent
  Transportation Systems (ITSC)}, 2018, pp. 602--607.

\bibitem{Hess2018}
D.~Heß, R.~Lattarulo, J.~Pérez, J.~Schindler, T.~Hesse, and F.~Köster,
  ``Fast maneuver planning for cooperative automated vehicles,'' in \emph{21st
  International Conference on Intelligent Transportation Systems (ITSC)}, 2018,
  pp. 1625--1632.

\bibitem{Schurmann2017}
B.~Schürmann, D.~Heß, J.~Eilbrecht, O.~Stursberg, F.~Köster, and M.~Althoff,
  ``Ensuring drivability of planned motions using formal methods,'' in
  \emph{IEEE 20th International Conference on Intelligent Transportation
  Systems (ITSC)}, 2017, pp. 1--8.

\bibitem{Guanetti2018}
J.~Guanetti, Y.~Kim, and F.~Borrelli, ``Control of connected and automated
  vehicles: State of the art and future challenges,'' \emph{Annual Reviews in
  Control}, vol.~45, pp. 18--40, 2018.

\bibitem{Trautman2010}
P.~Trautman and A.~Krause, ``Unfreezing the robot: Navigation in dense,
  interacting crowds,'' in \emph{IEEE/RSJ International Conference on
  Intelligent Robots and Systems}, 2010, pp. 797--803.

\bibitem{Buckman2019}
N.~Buckman, A.~Pierson, W.~Schwarting, S.~Karaman, and D.~Rus, ``Sharing is
  caring: Socially-compliant autonomous intersection negotiation,'' in
  \emph{IEEE/RSJ International Conference on Intelligent Robots and Systems
  (IROS)}, 2019, pp. 6136--6143.

\bibitem{Treiber2000}
M.~Treiber, A.~Hennecke, and D.~Helbing, ``Congested traffic states in
  empirical observations and microscopic simulations,'' \emph{Physical review.
  E, Statistical physics, plasmas, fluids, and related interdisciplinary
  topics}, vol. 62 2 Pt A, pp. 1805--24, 2000.

\bibitem{Sumo2018}
P.~A. Lopez, M.~Behrisch, L.~Bieker-Walz, J.~Erdmann, Y.-P. Fl{\"o}tter{\"o}d,
  R.~Hilbrich, L.~L{\"u}cken, J.~Rummel, P.~Wagner, and E.~Wie{\ss}ner,
  ``Microscopic traffic simulation using {SUMO},'' in \emph{21st IEEE
  International Conference on Intelligent Transportation Systems}.\hskip 1em
  plus 0.5em minus 0.4em\relax IEEE, 2018.

\bibitem{Acados2021}
R.~Verschueren, G.~Frison, D.~Kouzoupis, J.~Frey, N.~van Duijkeren, A.~Zanelli,
  B.~Novoselnik, T.~Albin, R.~Quirynen, and M.~Diehl, ``acados -- a modular
  open-source framework for fast embedded optimal control,'' \emph{Mathematical
  Programming Computation}, Oct 2021.

\end{thebibliography}
%\newpage

\begin{IEEEbiography}[{\includegraphics[width=1in,height=1.25in,clip,keepaspectratio]{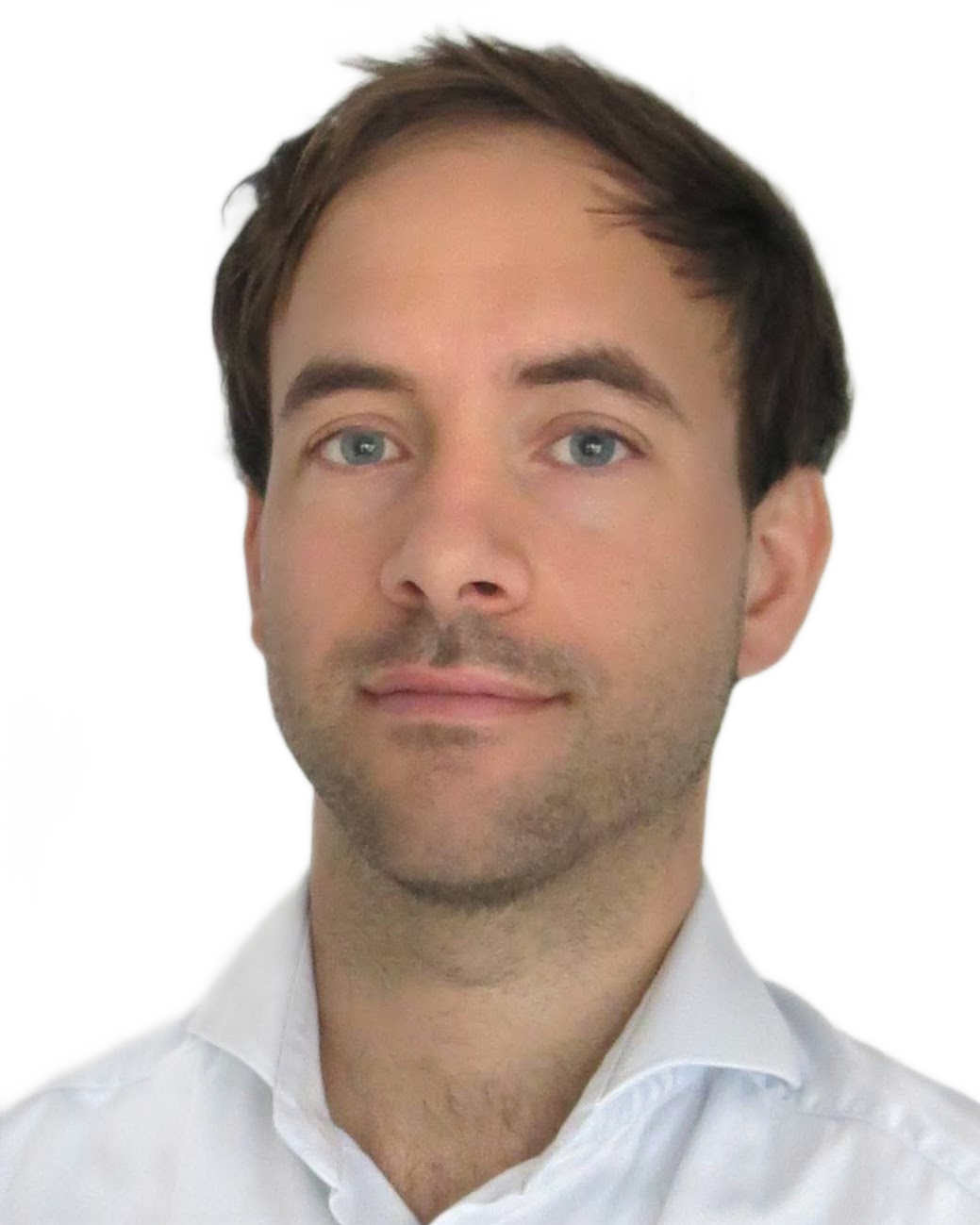}}]{Rudolf Reiter} received his Master's degree in 2016 in Electrical Engineering from the Graz University of Technology, Austria. From 2016 to 2018 he worked as a Control Systems Specialist at the Anton Paar GmbH, Graz, Austria. From 2018 to 2021 he worked as a researcher for the Virtual Vehicle Research Center, in Graz, Austria, where he started his Ph.D. in 2020 under the supervision of Prof. Dr. Mortiz Diehl. Since 2021, he continued his Ph.D. at a Marie-Skłodowska Curie Innovative Training Network position at the University of Freiburg, Germany. His research focus is within the field of learning- and optimization-based motion planning and control for autonomous vehicles and he is an active member of the Autonomous Racing Graz team.
\end{IEEEbiography}

\begin{IEEEbiography}[{\includegraphics[width=1in,height=1.25in,clip,keepaspectratio]{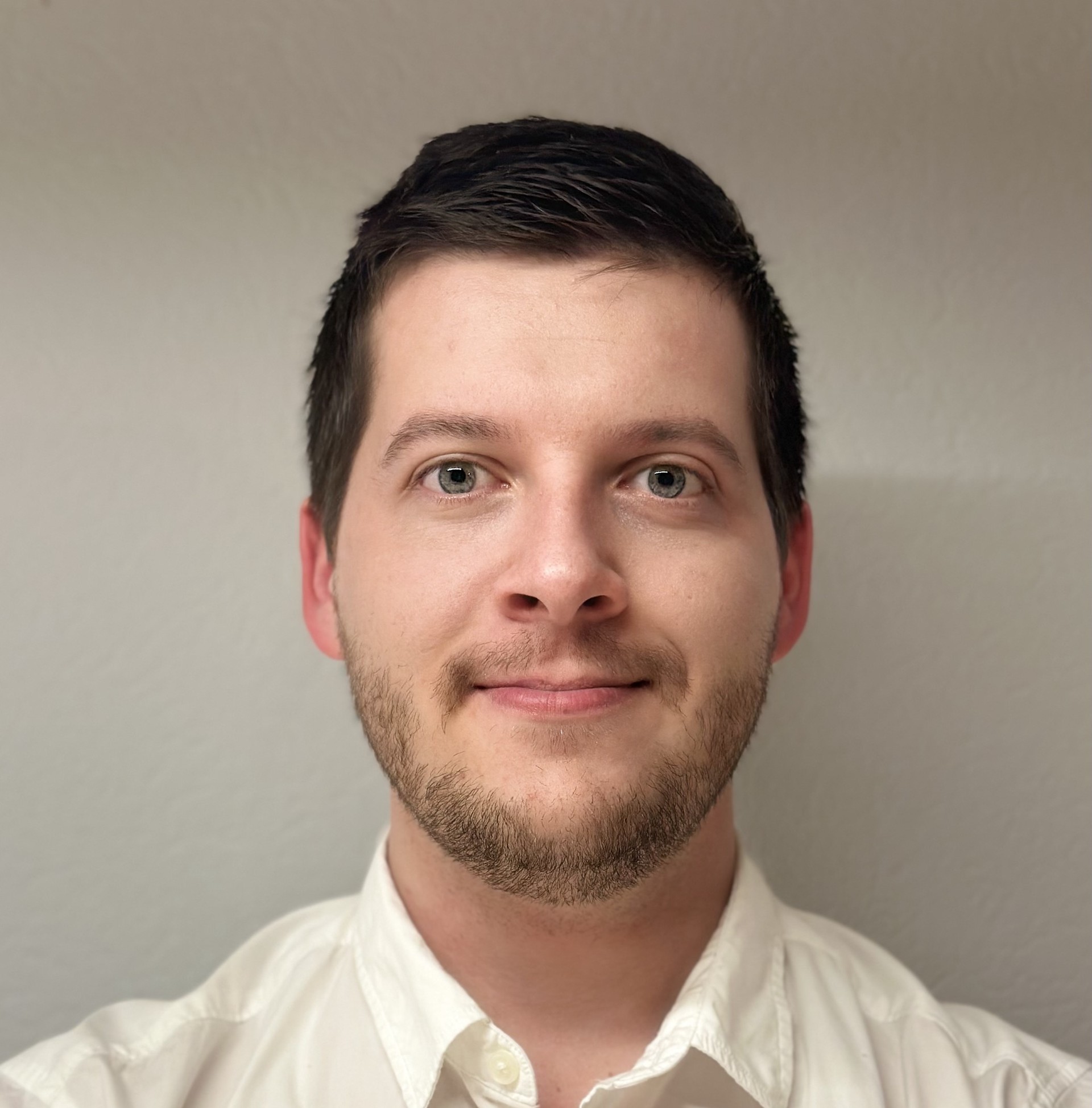}}]{Armin Nurkanovi\'c} received the B.Sc. degree from the Faculty of Electrical Engineering, Tuzla, Bosnia and Herzegovina, in 2015, and the M.Sc. degree from the Department of Electrical and Computer Engineering, Technical University of Munich, Munich, Germany, in 2018. 
He defended his Ph.D. in 2023 at the Systems Control and Optimization Laboratory, Department of Microsystems Engineering, University of Freiburg, Germany, under the supervision of Prof. Moritz Diehl.
He received the IEEE Control Systems Letters Outstanding Paper Award in 2022. 
His research interests are in the area of numerical methods for model predictive control, nonlinear optimization, and optimal control of nonsmooth and hybrid dynamical systems.
\end{IEEEbiography}

\begin{IEEEbiography}[{\includegraphics[width=1in,height=1.25in,clip,keepaspectratio]{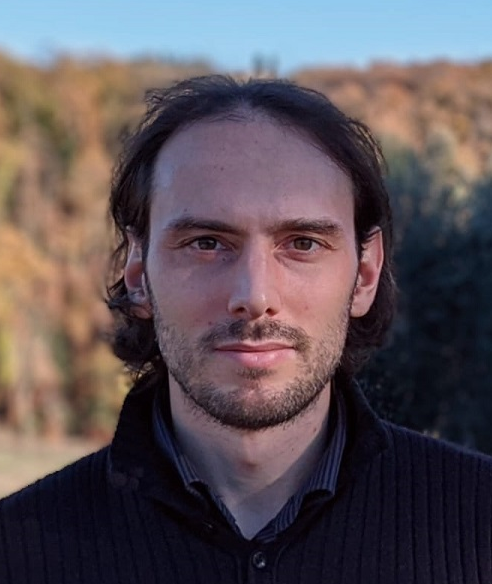}}]{Daniele Bernardini} received the Master’s degree in computer engineering in 2007 and the Ph.D. degree in information engineering  in 2011 from the University of
Siena, Italy, with a focus on model predictive control.
From 2011 to 2015 he was a Post-Doctoral
Fellow at IMT Lucca, Italy. In 2011
he co-founded ODYS S.r.l., where he is CTO, specializing in the development of efficient MPC solutions for embedded systems. He received the 2021 SAE Environmental Excellence in Transportation Award. His research interests include MPC, stochastic control, networked control systems, hybrid systems, and their application to real-time problems in the automotive, aerospace, robotics, and energy domains.
\end{IEEEbiography}

\begin{IEEEbiography}[{\includegraphics[width=1in,height=1.25in,clip,keepaspectratio]{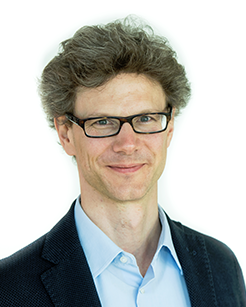}}]{Moritz Diehl}
	studied mathematics and physics at Heidelberg University, Germany and Cambridge University, U.K., and received the Ph.D. degree in optimization and nonlinear model predictive control from the Interdisciplinary Center for Scientific Computing, Heidelberg University, in 2001. From 2006 to 2013, he was a professor with the Department of Electrical Engineering, KU Leuven University Belgium. Since 2013, he is a professor at the University of Freiburg, Germany, where he heads the Systems Control and Optimization Laboratory, Department of Microsystems Engineering (IMTEK), and is also with the Department of Mathematics. His research interests include optimization and control, spanning from numerical method development to applications in different branches of engineering, with a focus on embedded and on renewable energy systems.
\end{IEEEbiography}

\begin{IEEEbiography}[{\includegraphics[width=1in,height=1.25in,clip,keepaspectratio]{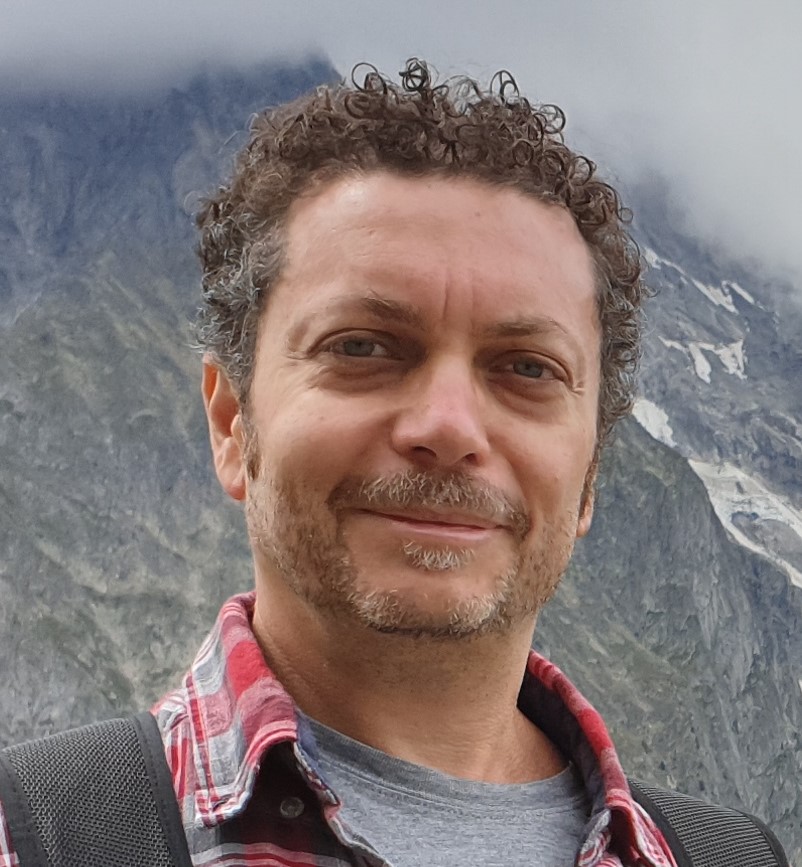}}]{Alberto Bemporad} (Fellow, IEEE) received his Master's degree cum laude in Electrical Engineering in 1993 and his Ph.D. in Control Engineering in 1997 from the University of Florence, Italy. In 1996/97 he was with the Center for Robotics and Automation, Department of Systems Science \& Mathematics, Washington University, St. Louis. In 1997-1999 he held a postdoctoral position at the Automatic Control Laboratory, ETH Zurich, Switzerland, where he collaborated as a Senior Researcher until 2002. In 1999-2009 he was with the Department of Information Engineering of the University of Siena, Italy, becoming an Associate Professor in 2005. In 2010-2011 he was with the Department of Mechanical and Structural Engineering of the University of Trento, Italy. Since 2011 he has been a Full Professor at the IMT School for Advanced Studies Lucca, Italy, where he served as the Director of the institute in 2012-2015. He spent visiting periods at Stanford University, University of Michigan, and Zhejiang University. In 2011 he co-founded ODYS S.r.l., a company specialized in developing model predictive control systems for industrial production. He has published more than 400 papers in the areas of model predictive control, hybrid systems, optimization, automotive control, and is the co-inventor of 21 patents. He is the author or coauthor of various software packages for model predictive control design and implementation, including the Model Predictive Control Toolbox (The Mathworks, Inc.) and the Hybrid Toolbox for MATLAB.
He was an Associate Editor of the IEEE Transactions on Automatic Control during 2001-2004 and Chair of the Technical Committee on Hybrid Systems of the IEEE Control Systems Society in 2002-2010. He received the IFAC High-Impact Paper Award for the 2011-14 triennial, the IEEE CSS Transition to Practice Award in 2019, and the 2021 SAE Environmental Excellence in Transportation Award. He has been an IEEE Fellow since 2010.
\end{IEEEbiography}
\end{document}